\documentclass[10pt,twocolumn,letterpaper]{article}

\usepackage{cvpr}
\usepackage{times}
\usepackage{epsfig}
\usepackage{graphicx}
\usepackage{amsmath}
\usepackage{amssymb}
\usepackage[utf8]{inputenc} 
\usepackage[T1]{fontenc}    
\usepackage{url}            
\usepackage{booktabs}       
\usepackage{amsfonts}       
\usepackage{nicefrac}       
\usepackage{microtype}      
\usepackage{gensymb}
\usepackage{dsfont}
\usepackage{wrapfig}

\usepackage{algorithm}
\usepackage[noend]{algpseudocode}
\usepackage{tabularx}
\usepackage{booktabs}
\usepackage{xcolor}
\usepackage{bbm}
\usepackage{color}
\usepackage{caption} 

\usepackage{enumitem}

\usepackage[pagebackref=true,breaklinks=true,letterpaper=true,colorlinks,bookmarks=false]{hyperref}

\cvprfinalcopy 

\ifcvprfinal\fi
\begin{document}

\title{Visual Reaction: Learning to Play Catch with Your Drone}

\author{Kuo-Hao Zeng$^{1}$\ \ \ \ \ \ Roozbeh Mottaghi$^{1,2}$\ \ \ \ \ \ Luca Weihs$^{2}$\ \ \ \ \ \ Ali Farhadi$^{1}$\\
\normalsize{$^{1}$Paul G. Allen School of Computer Science \& Engineering, University of Washington}\\
\normalsize{$^{2}$PRIOR @ Allen Institute for AI}
}

\maketitle
\thispagestyle{empty}

\begin{abstract}
In this paper we address the problem of \emph{visual reaction}: the task of interacting with dynamic environments where the changes in the environment are not necessarily caused by the agent itself. Visual reaction entails predicting the future changes in a visual environment and planning accordingly. We study the problem of visual reaction in the context of playing catch with a drone in visually rich synthetic environments. This is a challenging problem since the agent is required to learn (1) how objects with different physical properties and shapes move, (2) what sequence of actions should be taken according to the prediction, (3) how to adjust the actions based on the visual feedback from the dynamic environment (e.g., when objects bouncing off a wall), and (4) how to reason and act with an unexpected state change in a timely manner. We propose a new dataset for this task, which includes 30K throws of 20 types of objects in different directions with different forces. Our results show that our model that integrates a forecaster with a planner outperforms a set of strong baselines that are based on tracking as well as pure model-based and model-free RL baselines. The code and dataset are available at \url{github.com/KuoHaoZeng/Visual_Reaction}.

\end{abstract}

\vspace{-3mm}
\section{Introduction}

One of the key aspects of human cognition is the ability to interact and react in a visual environment. When we play tennis, we can predict how the ball moves and where it is supposed to hit the ground so we move the tennis racket accordingly. Or consider the scenario in which someone tosses the car keys in your direction and you quickly reposition your hands to catch them. These capabilities in humans start to develop during infancy and they are at the core of the cognition system \cite{bubic2010prediction,clark2013whatever}.

Visual reaction requires predicting the future followed by planning accordingly. The future prediction problem has received a lot of attention in the computer vision community. The work in this domain can be divided into two major categories. The first category considers predicting future actions of people or trajectories of cars (e.g., \cite{castrejon19,kitani12,lan14,walke17}). Typically, there are multiple correct solutions in these scenarios, and the outcome depends on the intention of the people. The second category is future prediction based on the physics of the scene (e.g., \cite{lerel16,mottaghi16b,watters17,zheng14}). The works in this category are mostly limited to learning from passive observation of images and videos, and there is no interaction or feedback involved during the prediction process. 

\begin{figure}[tp]
    \centering
    \includegraphics[width=20pc]{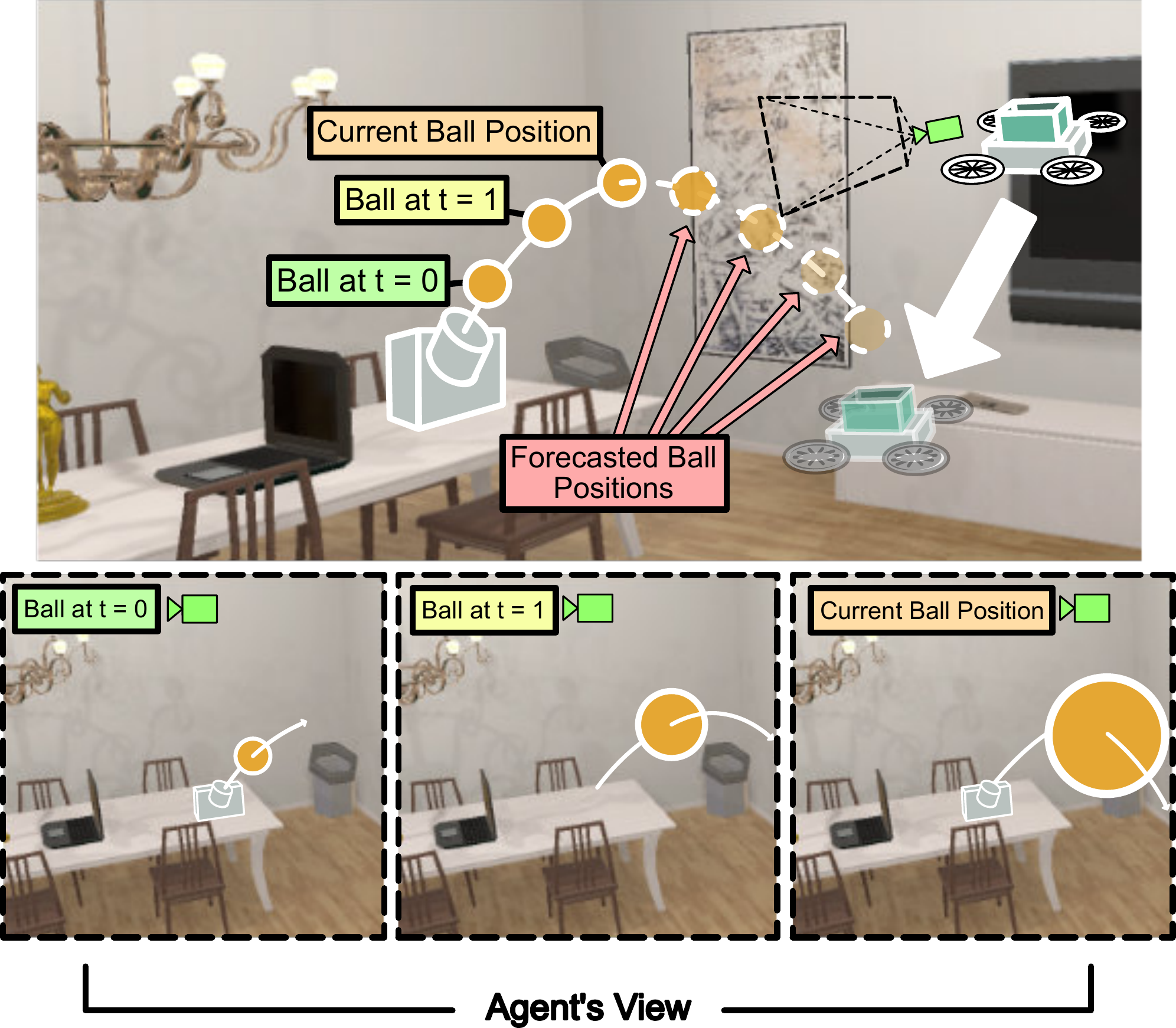}
    \caption{Our goal is to train an agent that can visually react with an interactive scene. In the studied task, the environment can evolve independently of the agent. There is a launcher in the scene that throws an object with different force magnitudes and in different angles. The drone learns to predict the trajectory of the object from ego-centric observations and move to a position that can catch the object. The trajectory of the thrown objects varies according to their weight and shape and also the magnitude and angle of the force used for throwing. }
    \label{fig:teaser}
    \vspace{-4mm}
\end{figure}

In this paper, we tackle the problem of \textit{visual reaction}: the task of predicting the future movements of objects in a dynamic environment and planning accordingly. The interaction enables us to make decisions on the fly and receive feedback from the environment to update our belief about the future movements. This is in contrast to passive approaches that perform prediction given pre-recorded images or videos. We study this problem in the context of playing catch with a drone, where the goal is to catch a thrown object using only visual ego-centric observations (Figure~\ref{fig:teaser}). Compared to the previous approaches, we not only need to predict future movements of the objects, but also to infer a minimal set of actions for the drone to catch the object in a timely manner. 

This problem exhibits various challenges. First, objects have different weights, shapes and materials, which makes their trajectories very different. Second, the trajectories vary based on the magnitude and angle of the force used for throwing. Third, the objects might collide with the wall or other structures in the scene, and suddenly change their trajectory. Fourth, the drone movements are not deterministic so the same action might result in different movements. Finally, the agent has limited time to reason and react to the dynamically evolving scene to catch the object before it hits the ground. 

Our proposed solution is an adaptation of the model-based Reinforcement Learning paradigm. More specifically, we propose a forecasting network that rolls out the future trajectory of the thrown object from visual observation. We integrate the forecasting network with a model-based planner to estimate the best sequence of drone actions for catching the object. The planner is able to roll out sequences of actions for the drone using the dynamics model and an action sampler to select the best action at each time step. In other words, we learn a policy using the rollout of both object and agent movements.

We perform our experiments in AI2-THOR \cite{ai2thor}, a near-photo-realistic interactive environment which models physics of objects and scenes (object weights, friction, collision, etc). Our experiments show that the proposed model outperforms baselines that are based on tracking (current state estimation as opposed to forecasting) and also pure model-free and model-based baselines. We provide an ablation study of our model and show how the performance varies with the number of rollouts and also the length of the planning horizon. Furthermore, we show how the model performs for object categories unseen during training. 

The contributions of the paper are as follows: (1) We investigate the problem of \emph{visual reaction} in an interactive,  dynamic, and visually rich environment. (2) We propose a new framework and dataset for visual reaction in the context of playing catch with a drone. (3) We propose a solution by integrating a planner and a forecaster and show it significantly outperforms a number of strong baselines. (4) We provide various analyses to better evaluate the models. 


\vspace{-3mm}
\section{Related Work}
\noindent \textbf{Future prediction \& Forecasting.} Various works explore future prediction and forecasting from visual data. Several authors consider the problem of predicting the future trajectories of objects from individual \cite{mottaghi16,pintea14,walker16,walker14,walker15,yuen10} and multiple sequential \cite{alahi16,kitani12,xie13} images. Unlike these works, we control an agent that interacts with the environment which causes its observation and viewpoint to change over time. A number of approaches explore prediction from ego-centric views. \cite{park16} predict a plausible set of ego-motion trajectories. \cite{rhinehart17} propose an Inverse Reinforcement Learning approach to predict the behavior of a person wearing a camera. \cite{vondrick16} learn visual representation from unlabelled video and use the representation for forecasting objects that appear in an ego-centric video. \cite{lee17} predict the future trajectories of interacting objects in a driving scenario. Our agent also forecasts the future trajectory based on ego-centric views of objects, but the prediction is based on physical laws (as opposed to peoples intentions). The problem of predicting future actions or the 3D pose of humans has been explored by \cite{chao17,fragkiadaki15,lan14,chen19}. Also, \cite{castrejon19,mathieu16,srivastava15,villegas19,villegas17,xue16} propose methods for generating future frames. Our task is different from the mentioned approaches as they use pre-recorded videos or images during training and inference, while we have an interactive setting. Methods such as \cite{finn16} and \cite{dosovitskiy17} consider future prediction in interactive settings. However, \cite{finn16} is based on a static third-person camera and \cite{dosovitskiy17} predicts the effect of agent actions and does not consider the physics of the scene.

\noindent \textbf{Planning.} There is a large body of work (e.g., \cite{chebotar17,gu16,hafner2018planet,heess15,nagabandi18,racaniere17,silver17,tamar16,wang18}) that involves a model-based planner. Our approach is similar to these approaches as we integrate the forecaster with a model-based planner. The work of  \cite{Buesing2018LearningAQ} shares similarities with our approach. The authors propose learning a compact latent state-space model of the environment and its dynamics; from this model an Imagination-Augmented Agent \cite{racaniere17} learns to produce informative rollouts in the latent space which improve its policy. We instead consider visually complex scenarios in 3D so learning a compact generative model is not as straightforward. Also, \cite{wang18} adopts a model-based planner for the task of vision and language navigation. They roll out the future states of the agent to form a model ensemble with model-free RL. Our task is quite different. Moreover, we consider the rollouts for both the agent and the moving object, which makes the problem more challenging.   

\noindent \textbf{Object catching in robotics.} The problem of catching objects has been studied in the robotics community. Quadrocopters have been used for juggling a ball \cite{muller11}, throwing and catching a ball \cite{ritz12}, playing table tennis \cite{silva15}, and catching a flying ball \cite{su17}. \cite{kim14} consider the problem of catching in-flight objects with uneven shapes. These approaches have one or multiple of the following issues: they use multiple external cameras and landmarks to localize the ball,  bypass the vision problem by attaching a distinctive marker to the ball, use the same environment for training and testing, or assume a stationary agent. We acknowledge that experiments on real robots involve complexities such as dealing with air resistance and mechanical constraints that are less accurately modeled in our setting. 

\noindent \textbf{Visual navigation.} There are various works that address the problem of visual navigation towards a static target using deep reinforcement learning or imitation learning (e.g., \cite{gupta17,mirowski17,savinov18,yang19,zhu17}). Our problem can be considered as an extension of these works since our target is moving and our agent has a limited amount of time to reach the target. Our work is also different from drone navigation (e.g., \cite{gandhi17,sadeghi17}) since we tackle the visual reaction problem.

\noindent \textbf{Object tracking.} Our approach is different from object tracking (e.g., \cite{bertinetto16,danelljan17,feichtenhofer17,nam16,sun18}) as we forecast the future object trajectories as opposed to the current location. Also, tracking methods typically provide only the location of the object of interest in video frames and do not provide any mechanism for an agent to take actions. 


\vspace{-3mm}
\section{Approach} \label{sec:methods}
We first define our task, \textit{visual reaction}: the task of interacting with dynamic environments that can evolve independently of the agent. Then, we provide an overview of the model. Finally, we describe each component of the model. 

\subsection{Task definition}
The goal is to learn a policy to catch a thrown object using an agent that moves in 3D space. There is a launcher in the environment that throws objects in the air with different forces in different directions. The agent needs to predict the future trajectory of the object from the past observations (three consecutive RGB images) and take actions at each timestep to intercept the object. An episode is successful if the agent catches the object, i.e. the object lies within the agent's top-mounted basket, before the object reaches the ground. The trajectories of objects vary depending on their physical properties (e.g., weight, shape, and material). The object might also collide with walls, structures, or other objects, and suddenly change its trajectory. 

For each episode, the agent and the launcher start at a random position in the environment (more details in Sec.~\ref{sec:framework}). The agent must act quickly to reach the object in a short time before the object hits the floor or goes to rest. This necessitates the use of a forecaster module that should be integrated with the policy of the agent. We consider 20 different object categories such as basketball, newspaper, and bowl (see Sec.~\ref{app:A} for the complete list). 

The model receives ego-centric RGB images from a camera that is mounted on top of the drone agent as input, and outputs an action $a_{d_t} =(\Delta_{v_x},\Delta_{v_y},\Delta_{v_z}) \in [-25 m/s^{2}, 25 m/s^{2}]^3$ for each timestep $t$, where, for example, $\Delta_{v_x}$ shows acceleration, in meters, along the $x$-axis. The movement of the agent is not deterministic due to the time dependent integration scheme of the physics engine. In the following, we denote the agent and object state by $s_{d} = [d, v_{d}, a_{d}, \phi, \theta]$ and $s_{o} = [o, v_{o}, a_{o}]$, respectively. $d$, $v_d$ and $a_d$ denote the position, velocity and acceleration of the drone and $o$, $v_o$ and $a_o$ denote those of the object. $\phi$ and $\theta$ specify the orientation of the agent camera, which can rotate independently from the agent. 



\subsection{Model Overview}
\begin{figure}[tp]
    \centering
    \includegraphics[width=19pc]{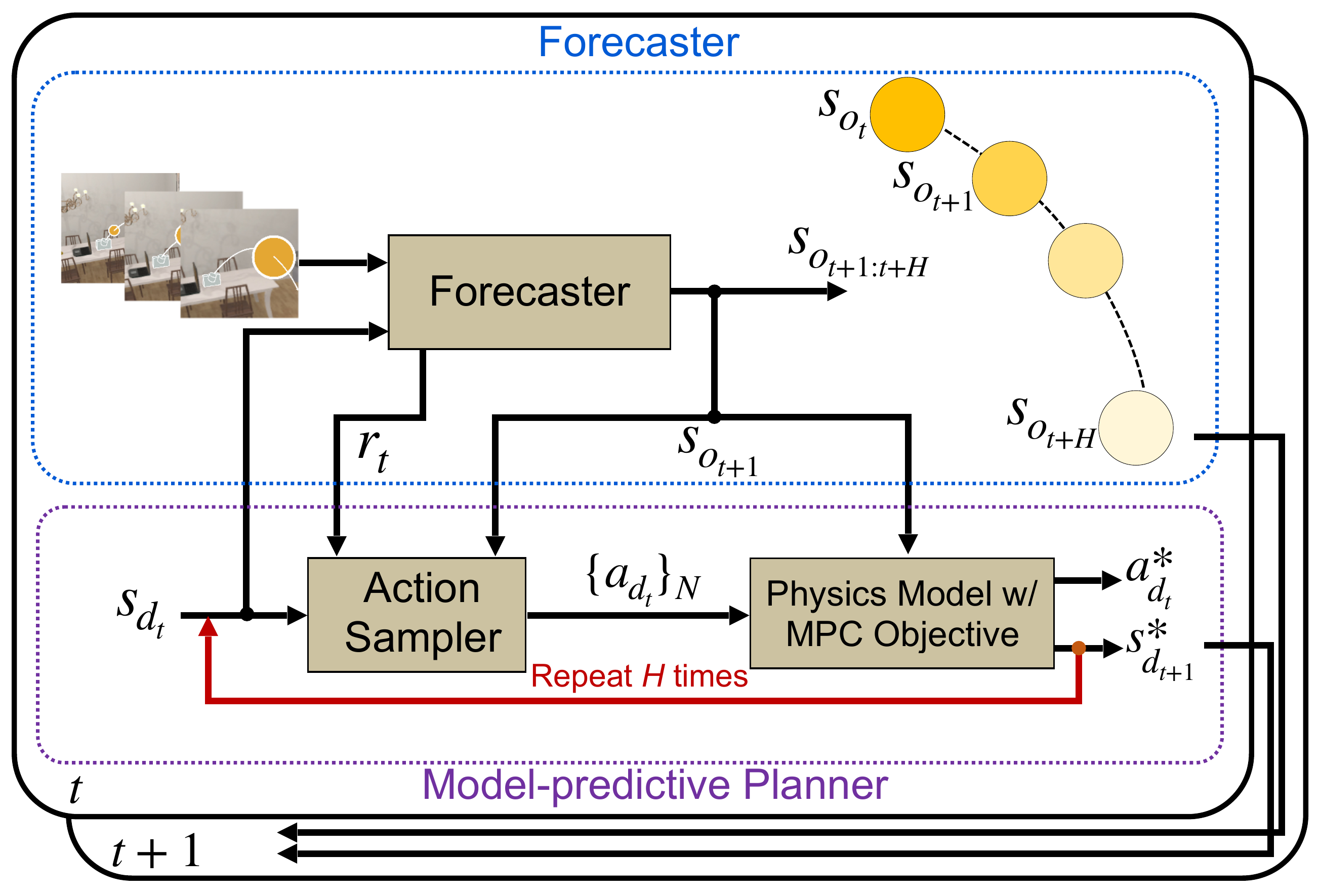}
    \caption{\textbf{Model overview.} Our model includes two main parts:  Forecaster and Planner. The visual encoding of the frames, object state, agent state and action are denoted by $r$, $s_{o}$, $s_{d}$, and $a$, respectively. $t$ denotes the timestep, and $H$ is the planning horizon.} 
    \label{fig:framework}
    \vspace{-6mm}
\end{figure}

Our model has two main components: a \textit{forecaster} and a \textit{model-predictive planner}, as illustrated in Fig.~\ref{fig:framework}. 
The forecaster receives the visual observations $i_{t-2:t}$ and the estimated agent state $s_{d_t}$ at time $t$, and predicts the current state $s_{o_t}$ of the thrown object. The forecaster further uses the predicted object state (i.e., position, velocity and acceleration) to forecast $H$ steps of object states $s_{o_{t+1:t+H}}$ in the future. The model-predictive planner is responsible for generating the best action for the agent such that it intercepts the thrown object. The model-predictive planner receives the future trajectory of the object from the forecaster and also the current estimate of the agent state as input and outputs the best action accordingly. The model-predictive planner includes an action sampler whose goal is to sample $N$ sequences of actions given the current estimate of the agent state, the predicted object trajectory, and the intermediate representation $r_t$ produced by the visual encoder in the forecaster. The action sampler samples actions according to a policy network that is learned. The second component of the model-predictive planner consists of a physics model and a model-predictive controller (MPC). The physics model follows Newton Motion Equation to estimate the next state of the agent (i.e., position and velocity at the next timestep) given the current state and action (that is generated by the action sampler). Our approach builds on related joint model-based and model-free RL ideas. However, instead of an ensemble of model-free and model-based RL for better decision making \cite{kurutach2018model, wang18}, or using the dynamics model as a data augmentor/imaginer \cite{feinberg2018model, racaniere17} to help the training of model-free RL, we explicitly employ model-free RL to train an action sampler for the model-predictive planner.



\begin{figure*}[t!]
    \centering
    \includegraphics[width=41pc]{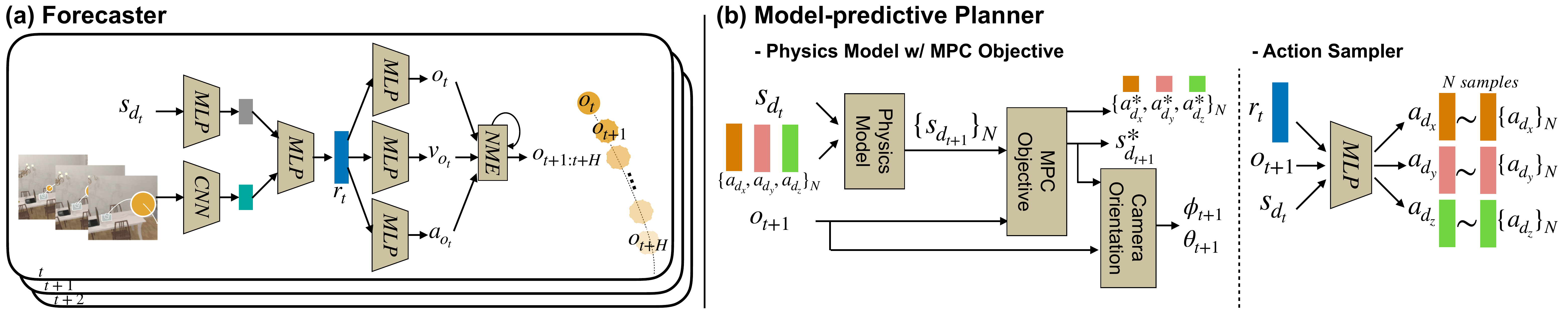}
    \caption{\textbf{Model architecture.} (a) The forecaster receives images and an estimate of the agent state $s_{d_t}$ as input and outputs the estimates for the current state $s_{o_t}$, including $o_t$, $v_{o_t}$, and $a_{o_t}$. Then it forecasts future positions of the object $o_{t+1:t+H}$ by discretized Newton Motion Equation. Forecasting is repeated every timestep if the object has not been caught. (b) The model-predictive planner includes a MPC w/ Physics model and an action sampler. The action sampler generates $N$ sequences $\textbf{a}_{d_{t:t+H-1}} = \{(\Delta^j_{v_{x,i}},\Delta^j_{v_{y,i}},\Delta^j_{v_{z,i}})_{i=t}^{t+H-1}| j=1,...,N \}$ of actions at each timestep, and an optimal action $(\Delta_{v_{x}^*},\Delta_{v_{y}^*},\Delta_{v_{z}^*})$ is chosen such that it minimizes the distance between the agent and the object at each timestep.}
    \label{fig:model} 
    \vspace{-4mm}
\end{figure*}

In the following, we begin by introducing our forecaster, as shown in Fig.~\ref{fig:model}(a), along with its training strategy. We then describe how we integrate the forecaster with the model-predictive planner, as presented in Fig.~\ref{fig:framework} and Fig.~\ref{fig:model}(b). Finally, we explain how we utilize model-free RL to learn the action distribution used in our planner, Fig.~\ref{fig:model}(b).


\subsection{Forecaster}\label{sec:fn}

The purpose of the forecaster is to predict the current object state $s_{o_t}$, which includes the position $o_t\in\mathbb{R}^3$, the velocity $v_{o_t}\in\mathbb{R}^3$, and the acceleration $a_{o_t}\in\mathbb{R}^3$, and then, based on the prediction, forecast future object positions $o_{t+1:t+H}$ from the most recent three consecutive images $i_{t-2:t}$. The reason for forecasting $H$ timesteps in the future is to enable the planner to employ MPC to select the best action for the task. We show how the horizon length $H$ affects the performance in Sec.~\ref{app:H}. Note that if the agent does not catch the object in the next timestep, we query the forecaster again to predict the trajectory of the object $o_{t+2:t+H+1}$ for the next $H$ steps. Forecaster also produces the intermediate visual representation $r_t\in\mathbb{R}^{256}$, which is used by the action sampler. The details are illustrated in Fig.~\ref{fig:model}(a). We define the positions, velocity, and acceleration in the agent's coordinate frame at its starting position. 


The three consecutive frames $i_{t-2:t}$ are passed through a convolutional neural network (CNN). The features of the images and the current estimate of the agent state $s_{d_t}$ are combined using an MLP, resulting in an embedding $r_t$. Then, the current state of the object $s_{o_t}$ is obtained from $r_t$ through three separate MLPs. The NME, which follows the discretized Newton's Motion Equation ($o_{t+1} = o_t + v_t\times\Delta t$, $v_{t+1} = v_{t} + a_{t}\times\Delta t$) receives the predicted state of the object to calculate the future positions $o_{t+1:t+H}$. We take the derivative of NME and back-propagate the gradients through it in the training phase. Note that NME itself is not learned. 

To train the forecaster, we provide the ground truth positions of the thrown object from the environment and obtain the velocity and acceleration by taking the derivative of the positions. We cast the position, velocity, and acceleration prediction as a regression problem and use the L1 loss for optimization.

\subsection{Model-predictive Planner}\label{sec:ap}
Given the forecasted trajectory of the thrown object, our goal is to control the flying agent to catch the object. We integrate the model-predictive planner with model-free RL to explicitly incorporate the output of the forecaster.


Our proposed model-predictive planner consists of a model-predictive controller (MPC) with a physics model, and an action sampler as illustrated in Fig.~\ref{fig:model}(b). We will describe how we design the action sampler in Sec.~\ref{sec:as}. The action sampler produces a rollout of future actions. The action is defined as the acceleration $a_{d}$ of the agent. We sample $N$ sequences of actions that are of length $H$ from the action distribution. We denote these $N$ sequences by $\textbf{a}_{d_{t:t+H-1}}$. For each action in the $N$ sequences, the physics model estimates the next state of the agent $s_{d_{t+1}}$ given the current state $s_{d_t}$ by using the discretized Newton's Motion Equation ($d_{t+1} = d_t + v_{d_t}\times\Delta t$, $v_{d_{t+1}} = v_{d_t} + a_{d_t}\times\Delta t$). This results in $N$ possible trajectories $d_{t+1:t+H}$ for the agent. Given the forecasted object trajectories $o_{t+1:t+H}$, the MPC then selects the best sequence of actions $\textbf{a}^{*}_{t:t+H-1}$ based on the defined objective. The objective for MPC is to select a sequence of actions that minimizes the sum of the distances between the agent and the object over $H$ timesteps. We select the first action $a^{*}_{t}$ in the sequence of actions, and the agent executes this action. We feed in the agent's next state $s^{*}_{d_{t+1}}$ for planning in the next timestep. 

\noindent \textbf{Active camera viewpoint.} The agent is equipped with a camera that rotates. The angle of the camera is denoted by $\phi$ and $\theta$ in the agent's state vector $s_d$. We use the estimated object and agent position at time $t+1$, $o_{t+1}$ and  $d^{*}_{t+1}$, to compute the angle of the camera. We calculate the relative position $p \in (p_x, p_y, p_z)$ between object and agent by $o - d$. Then, we obtain the Euler angles along $y$-axis and $x$-axis by $\arctan{\frac{p_x}{p_z}}$ and $\arctan{\frac{p_y}{p_z}}$, respectively. In Sec.~\ref{app:B}, we also show results for the case that the camera is fixed.

\subsection{Action sampler}\label{sec:as}
The actions can be sampled from a uniform distribution over the action space or a learned policy network. We take the latter approach and train a policy network which is conditioned on the forecasted object state, current agent state and visual representation. Model-based approaches need to sample a large set of actions at each timestep to achieve a high level of performance. To alleviate this issue, we parameterize our action sampler by a series of MLPs that learns an action distribution given the current agent state, the forecasted trajectory of the object $o_{t+1:t+H}$ and the visual representation $r_t$ of observation $i_t-2:t$ (refer to Sec.~\ref{sec:fn}). This helps to better shape the action distribution, which may result in requiring fewer samples and better performance. 


To train our policy network, we utilize policy gradients with the actor-critic algorithm \cite{sutton2000policy}. To provide the reward signal for the policy gradient, we use the `success' signal (if the agent catches the object or not) as a reward. In practice, if the agent succeeds to catch the object before it hits the ground or goes to rest, it would receive a reward of $+1$. Furthermore, we also measure the distance between the agent trajectory and the object trajectory as an additional reward signal (pointwise distance at each timestep). As a result, the total reward for each episode is $R=\mathbbm{1}\{\text{episode success}\}-0.01\cdot \sum_{t} ||d^*_{t}-o^*_t||_2$ where $d^*_{t}$ and $o^*_t$ are the ground truth positions of the agent and object at time $t$. 

\section{Experiments}

\begin{figure*}[tp]
    \centering
    \includegraphics[width=40pc,height=12pc]{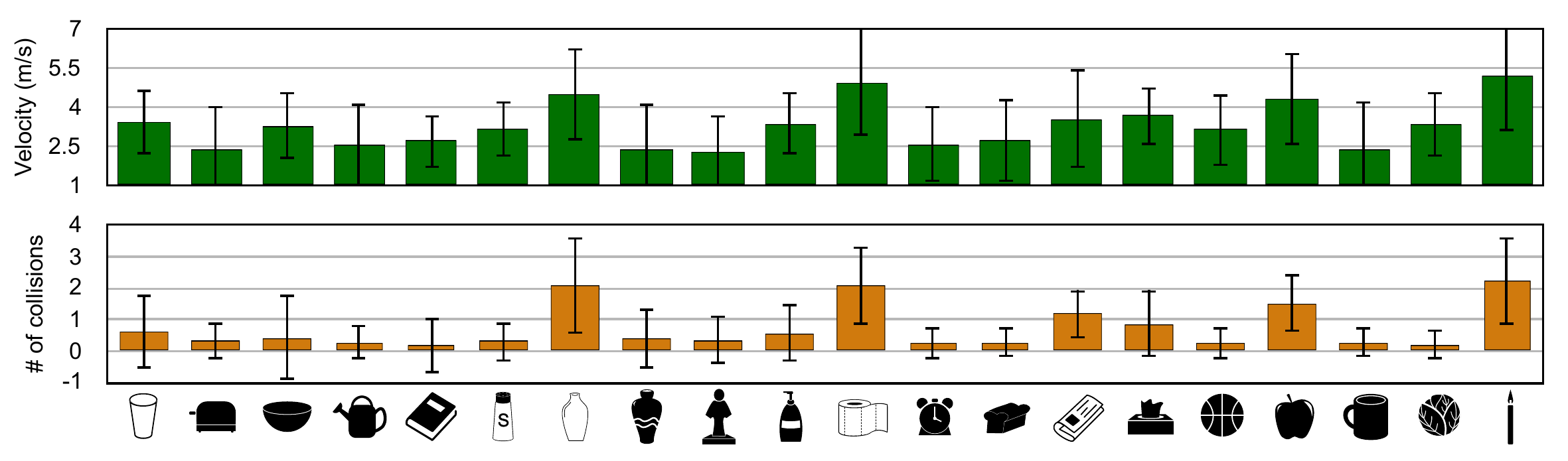} 
    \vspace{-2mm}
    \caption{\textbf{Dataset statistics.} We provide the statistics for the 20 types of objects in our dataset. We illustrate the average velocity along the trajectories and the number of collisions with walls or other structures in the scene.}
    \label{fig:stats}
    \vspace{-4mm}
\end{figure*}

We first describe the environment that we use for training and evaluating our model. We then provide results for a set of  baselines: different variations of using current state prediction instead of future forecasting and a model-free baseline. We also provide ablation results for our method, where we use uniform sampling instead of the learned action sampler. Moreover, we study how the performance changes with varying mobility of the agent, noise in the agent movement and number of action sequence samples. Finally, we provide analysis of the results for each object category, different levels of difficulty, and objects unseen during training.

\subsection{Framework}
\label{sec:framework}
We use AI2-THOR \cite{ai2thor}, which is an interactive 3D indoor virtual environment with near photo-realistic scenes. We use AI2-THOR v2.3.8, which implements physical properties such as object materials, elasticity of various materials, object mass and includes a drone agent. We add a launcher to the scenes that throws objects with random force magnitudes in random directions. 

The trajectories of the objects vary according to their mass, shape, and material. Sometimes the objects collide with walls or other objects in the scene, which causes sudden changes in the trajectory. Therefore, standard equations of motion are not sufficient to estimate the trajectories, and learning from visual data is necessary. The statistics of the average velocity  and the number of collisions have been provided in Fig.~\ref{fig:stats}. More information about the physical properties of the objects are in Sec.~\ref{app:C}.

The drone has a box on top to catch objects. The size of drone is $0.47m\times0.37m$ with a height of $0.14m$, and the box is $0.3m\times0.3m$ with a height of $0.2m$. The drone is equipped with a camera that is able to rotate. The maximum acceleration of the drone is $25 m/s^{2}$ and the maximum velocity is $40m/s$. However, we provide results for different maximum acceleration of the drone. The action for the drone is specified by acceleration in $x$, $y$, and $z$ directions. The action space is continuous, but is capped by the maximum acceleration and velocity.

\noindent \textbf{Experiment settings.} We use the \emph{living room} scenes of AI2-THOR for our experiments (30 scenes in total). We follow the common practice for AI2-THOR wherein the first 20 scenes are used for training, the next 5 for validation, and the last 5 for testing. The drone and the launcher are assigned a random position at the beginning of every episode. We set the horizontal relative distance between the launcher and the drone to be $2$ meters (any random position). We set the height of the launcher to be $1.8$ meters from the ground which is similar to the average human height. The drone faces the launcher in the beginning of each episode so it observes that an object is being thrown. 

To throw the object, the launcher randomly selects a force between $[40, 60]$ newtons, an elevation angle between $[45, 60]$ degree, and an azimuth angle between $[-30, 30]$ degree for each episode. The only input to our model at inference time is the ego-centric RGB image from the drone. We use 20 categories of objects such as basketball, alarm clock, and apple for our experiments. We observe different types of trajectories such as parabolic motion, bouncing off the walls and collision with other objects, resulting in sharp changes in the direction. Note that each object category has different physical properties (mass, bounciness, etc.) so the trajectories are quite different. We use the same objects for training and testing. However, the scenes, the positions, the magnitude, and the angle of the throws vary at test time. We also show an experiment, where we test the model on categories unseen during training. We consider 20K trajectories during training, 5K for val and 5K for test. The number of trajectories is uniform across all object categories. 

\subsection{Implementation details}\label{sec:TEdetail}
We train our model by first training the forecaster. Then we freeze the parameters of the forecaster, and train the action sampler. An episode is successful if the agent catches the object. We end an episode if the agent succeeds in catching the object, the object falls on the ground, or the length of the episode exceeds $50$ steps which is equal to $1$ second. We use SGD with initial learning rate of $10^{-1}$ for forecaster training and decrease it by a factor of $10$ every $1.5\times10^{4}$ iterations. For the policy network, we employ Adam optimizer \cite{kingma2014adam} with a learning rate of $10^{-4}$. We evaluate the framework every $10^{3}$ iterations on the validation scenes and stop the training when the success rate saturates. We use MobileNet v2~\cite{sandler2018mobilenetv2}, which is an efficient and light-weight network as our CNN model. The forecaster outputs the current object position, velocity, and acceleration. The action sampler provides a set of accelerations to the planner. They are all continuous numbers. Sec.~\ref{app:D} provides details for the architecture of each component of the model.

\begin{table*}[tp]
    \vspace{-0.4cm}
    \setlength\extrarowheight{1pt}
    \setlength{\tabcolsep}{3pt}
	\centering
	\begin{tabular}{c || c | c | c | c | c | c |}
	    \cline{2-7}
	    & \textbf{N = 100000}  & \textbf{N = 10000} & \textbf{N = 1000} & \textbf{N = 100} & \textbf{N = 10}   & \textbf{Best} \\ 
	    \hline
	    \multicolumn{1}{|c||}{\textbf{Curr. Pos. Predictor (CPP)}}&  22.92$\pm{2.3}$     &  22.57$\pm{2.0}$    &   21.04$\pm{1.2}$   & 18.72$\pm{1.8}$     & 10.86$\pm{0.5}$  & 22.92$\pm{2.3}$  \\
	    \hline
        \multicolumn{1}{|c||}{\textbf{CPP + Kalman Filter}}&  23.22$\pm{1.29}$ &  22.78$\pm{0.90}$  & 21.88$\pm{0.79}$ &  19.29$\pm{0.81}$ &  12.17$\pm{1.2}$  &  23.22$\pm{1.29}$\\ \hline
        \multicolumn{1}{|c||}{\textbf{Model-free (A3C \cite{mnih2016asynchronous})}}&  -    &   -   &  -    &  -    &  -  & 4.54$\pm{2.3}$ \\ \hline
        \hline
        \multicolumn{1}{|c||}{\textbf{Ours, ME, uniform AS}}& 6.12$\pm{0.7}$ & 6.11$\pm{0.7}$ & 6.00$\pm{0.5}$ & 5.99$\pm{0.5}$ & 5.12$\pm{1.0}$ & 6.12$\pm{0.7}$\\ \hline
        \multicolumn{1}{|c||}{\textbf{Ours, uniform AS}}& 26.01$\pm{1.3}$ & 25.47$\pm{1.3}$ & 23.61$\pm{1.5}$ & 20.65$\pm{0.93}$ & 10.58$\pm{1.1}$ & 26.01$\pm{1.3}$\\ \hline
        \multicolumn{1}{|c||}{\textbf{Ours, full}}& \textbf{29.34$\pm{0.9}$} & \textbf{29.26$\pm{1.4}$} & \textbf{29.12$\pm{0.8}$} & \textbf{29.14$\pm{0.8}$} & \textbf{24.72$\pm{1.6}$}  & \textbf{29.34$\pm{0.9}$}  \\ \hline\hline
        \multicolumn{1}{|c||}{\textbf{MPC Upper bound}}& 68.67$\pm{1.9}$ & 76.00$\pm{0.0}$ & 78.67$\pm{1.9}$ & 66.00$\pm{3.3}$ & 49.33$\pm{10.5}$  & 78.67$\pm{1.9}$  \\ \hline
	\end{tabular}
	\vspace{-1mm}
	\caption{\textbf{Quantitative results.} We report the success rate for the baselines and the ablations of our model. $N$ refers to the number of action sequences that the action sampler provides. The model-free baseline does not have an action sequence sampling component so we can provide only one number. The MPC upper bound is the case that model-predictive planner uses perfect forecasting with uniform action sampler. Note that the MPC upper bound must be done in the off-line mode since the perfect forecasting only available after collecting the objects' trajectory.}
	\label{tab:result_mo}
	\vspace{-2mm}
\end{table*}

\begin{table*}[tp]
    \setlength\extrarowheight{1pt}
    \setlength{\tabcolsep}{3pt}
	\centering
	\hspace{-20mm}
	\resizebox{0.89\textwidth}{!}{\begin{minipage}{\textwidth}
	\begin{tabular}{c | c | c | c | c | c | c | c | c | c | c | c | c | c | c | c | c | c | c | c | c |}
	    \cline{2-21}
	    & \begin{minipage}{.03\textwidth}\centering \vspace{.1em}
      \includegraphics[width=5mm, height=5mm]{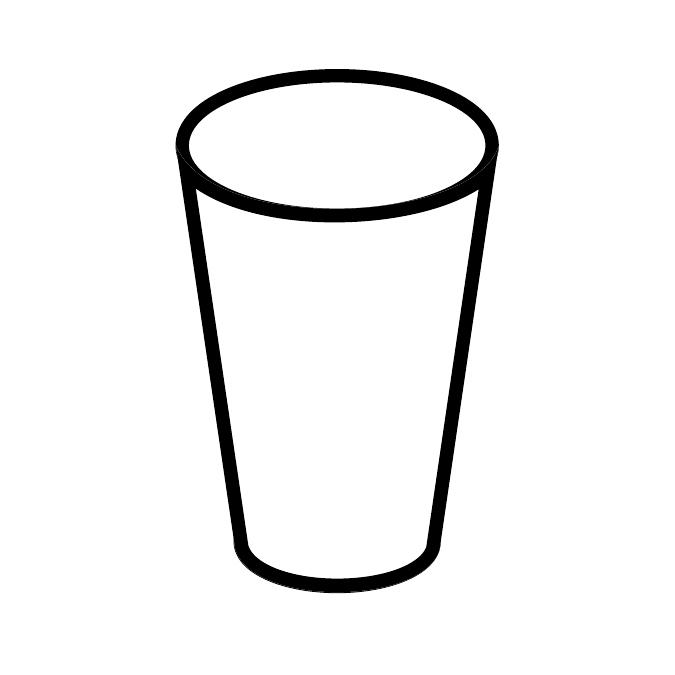} \end{minipage}  
        & \begin{minipage}{.03\textwidth}\centering \vspace{.1em}
      \includegraphics[width=5mm, height=5mm]{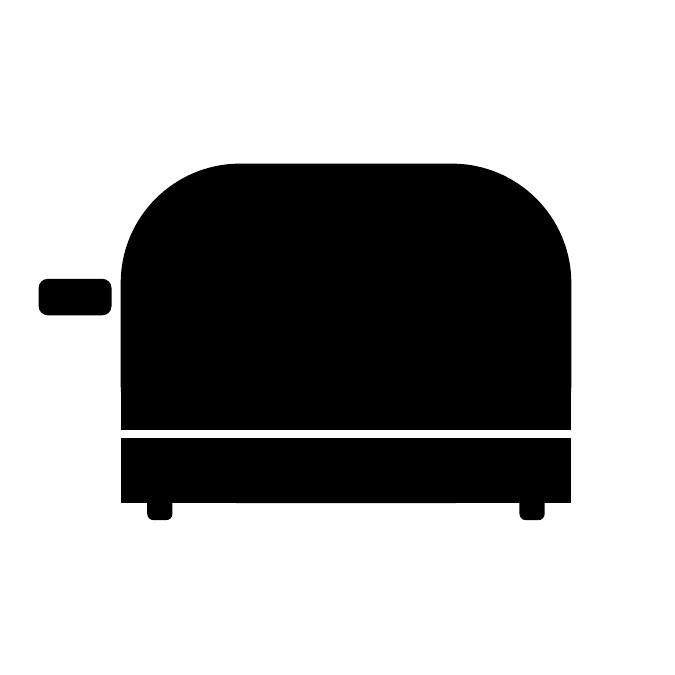} \end{minipage} 
        & \begin{minipage}{.03\textwidth}\centering \vspace{.1em}
      \includegraphics[width=5mm, height=5mm]{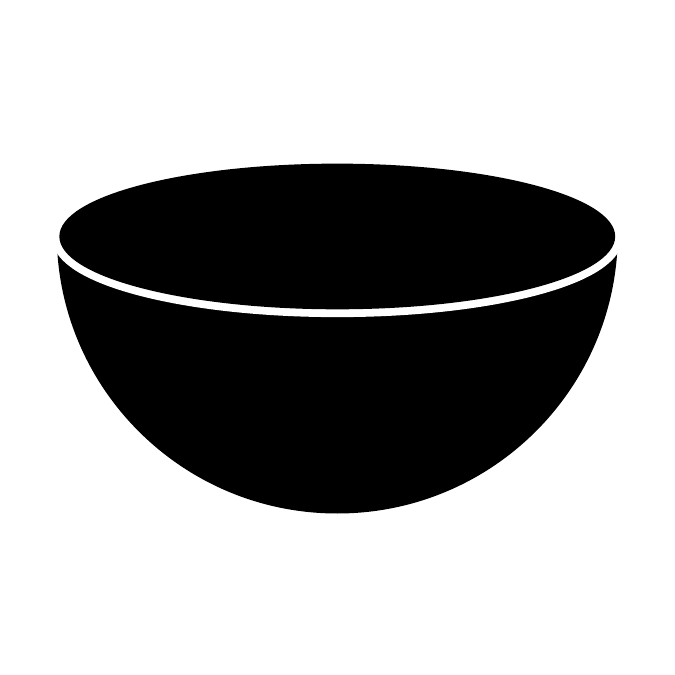} \end{minipage} 
        & \begin{minipage}{.03\textwidth}\centering \vspace{.1em}
      \includegraphics[width=5mm, height=5mm]{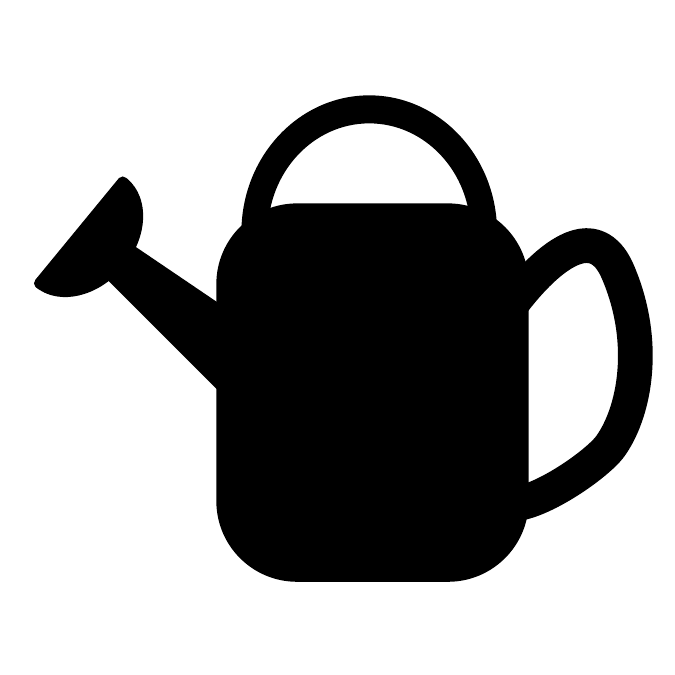} \end{minipage}
        & \begin{minipage}{.03\textwidth}\centering \vspace{.1em}
      \includegraphics[width=5mm, height=5mm]{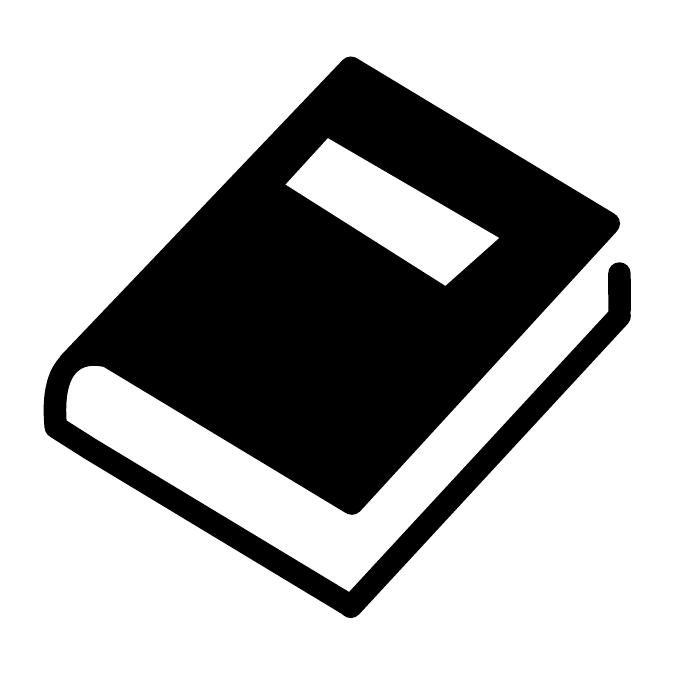} \end{minipage}
        & \begin{minipage}{.03\textwidth}\centering \vspace{.1em}
      \includegraphics[width=5mm, height=5mm]{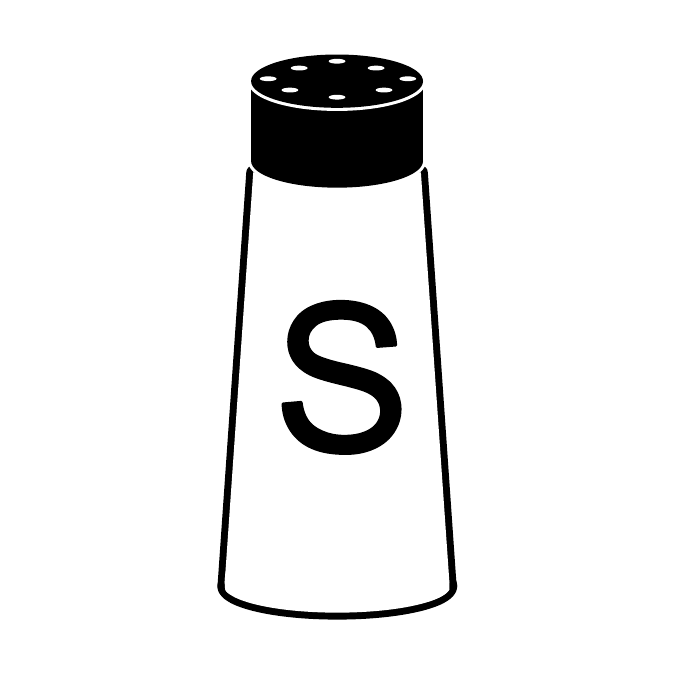} \end{minipage}
        & \begin{minipage}{.03\textwidth}\centering \vspace{.1em}
      \includegraphics[width=5mm, height=5mm]{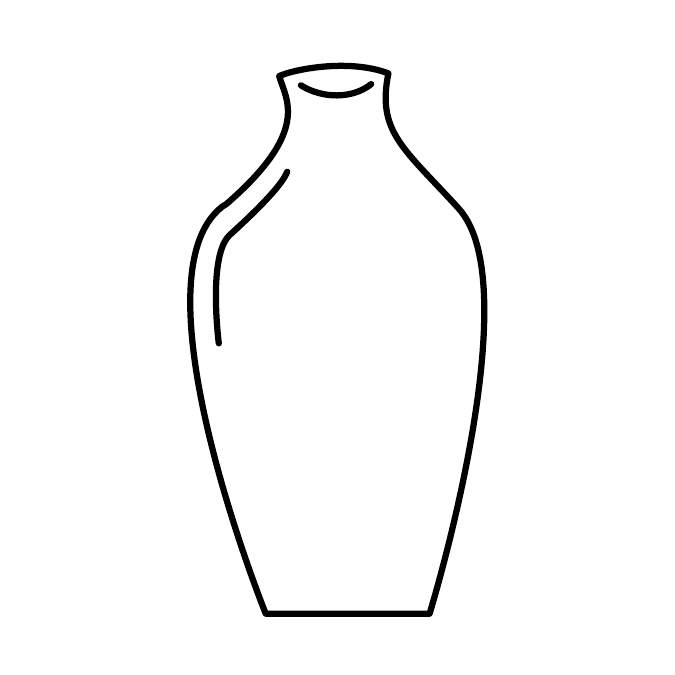} \end{minipage}
        & \begin{minipage}{.03\textwidth}\centering \vspace{.1em}
      \includegraphics[width=5mm, height=5mm]{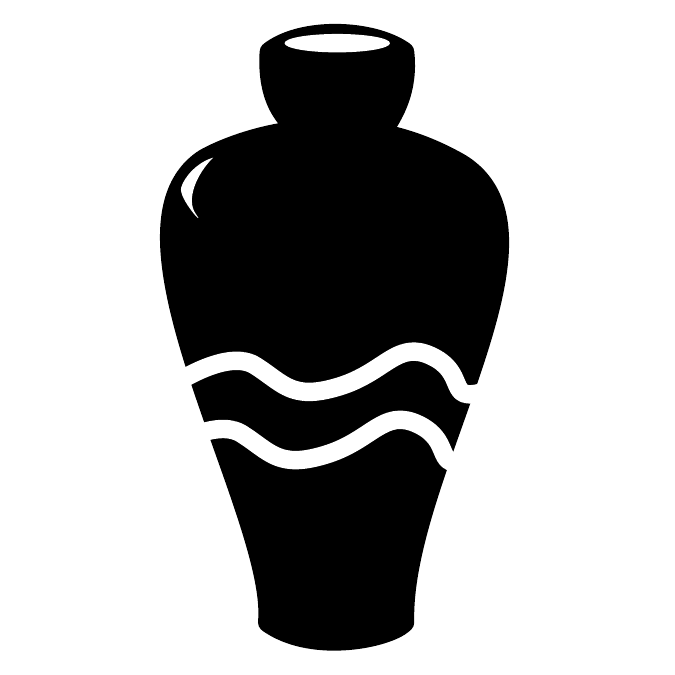} \end{minipage}
        & \begin{minipage}{.03\textwidth}\centering \vspace{.1em}
      \includegraphics[width=5mm, height=5mm]{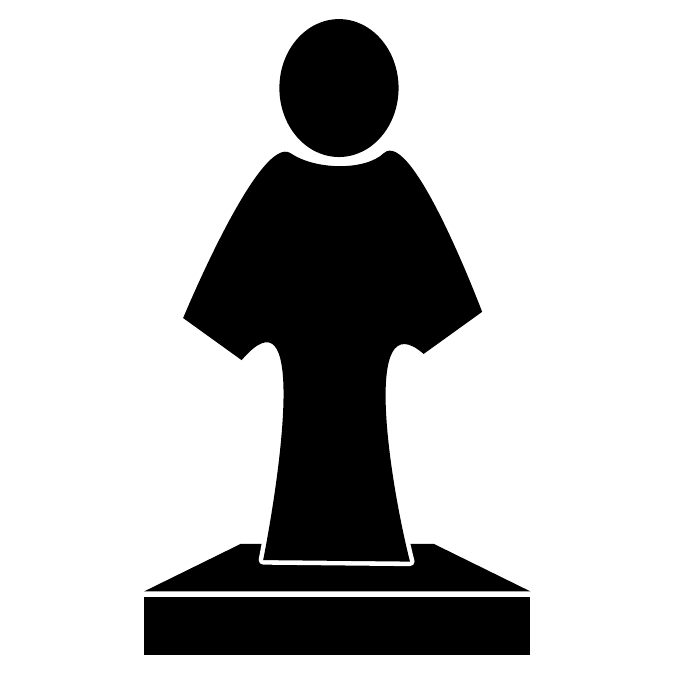} \end{minipage}
        & \begin{minipage}{.03\textwidth}\centering \vspace{.1em}
      \includegraphics[width=5mm, height=5mm]{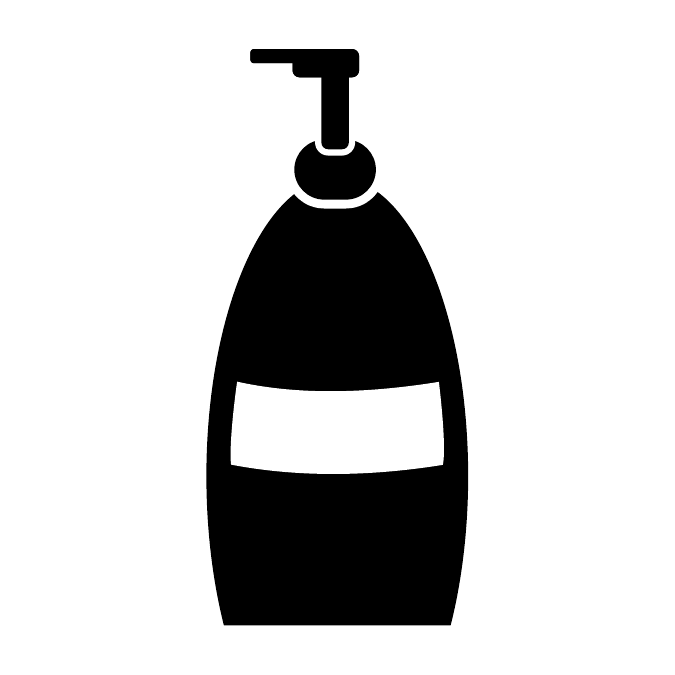} \end{minipage} 
      & \begin{minipage}{.03\textwidth}\centering \vspace{.1em}
      \includegraphics[width=5mm, height=5mm]{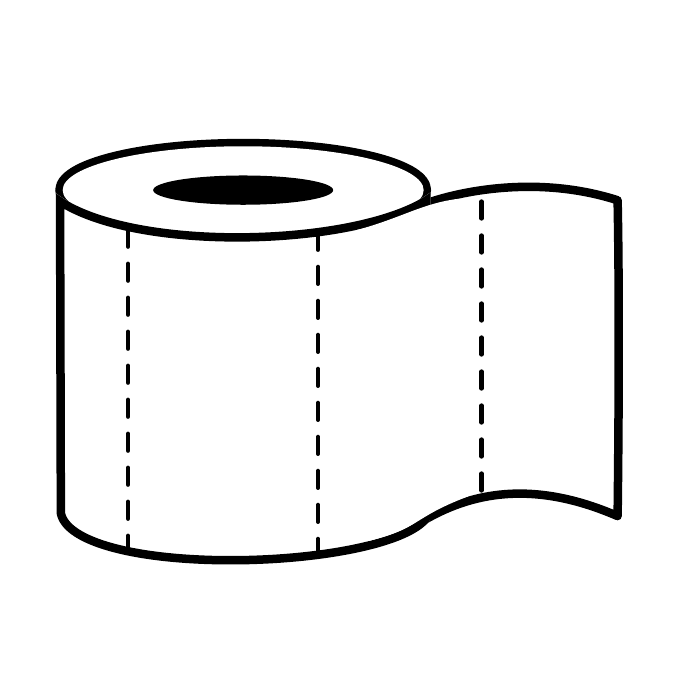} \end{minipage}
        & \begin{minipage}{.03\textwidth}\centering \vspace{.1em}
      \includegraphics[width=5mm, height=5mm]{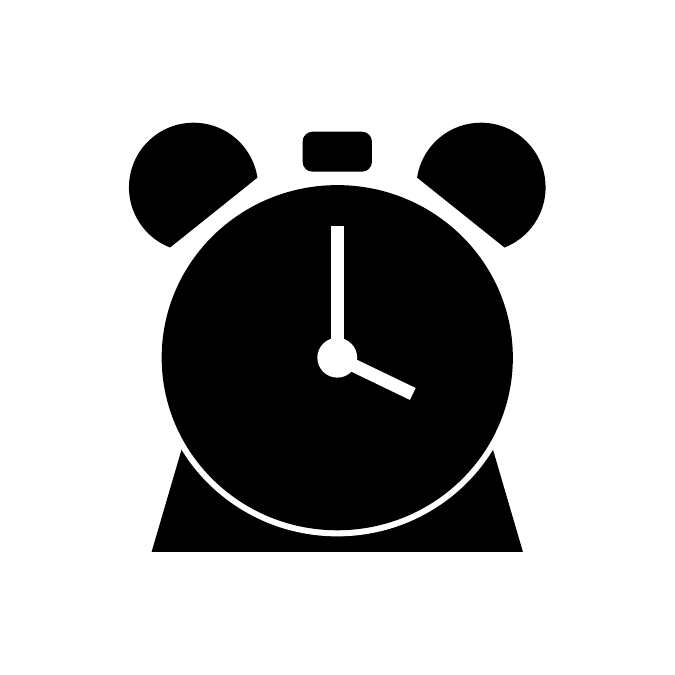} \end{minipage}
        & \begin{minipage}{.03\textwidth}\centering \vspace{.1em}
      \includegraphics[width=5mm, height=5mm]{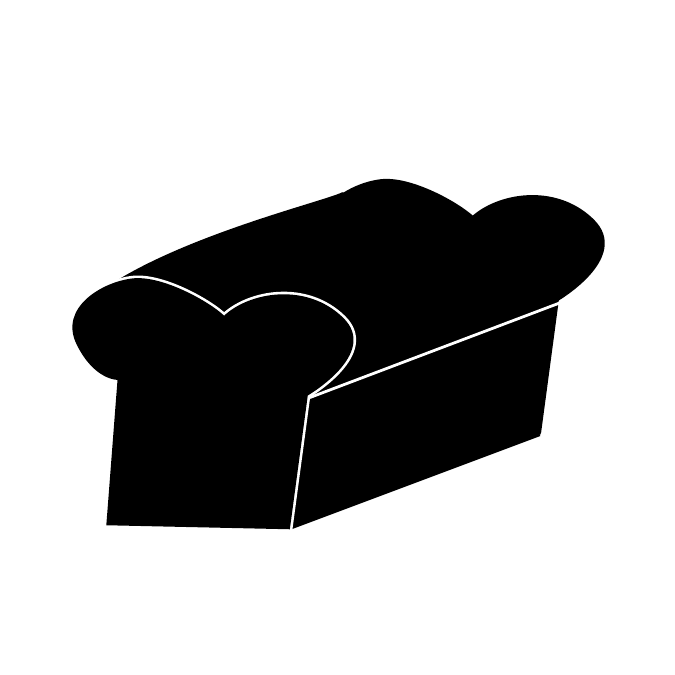} \end{minipage}
        & \begin{minipage}{.03\textwidth}\centering \vspace{.1em}
      \includegraphics[width=5mm, height=5mm]{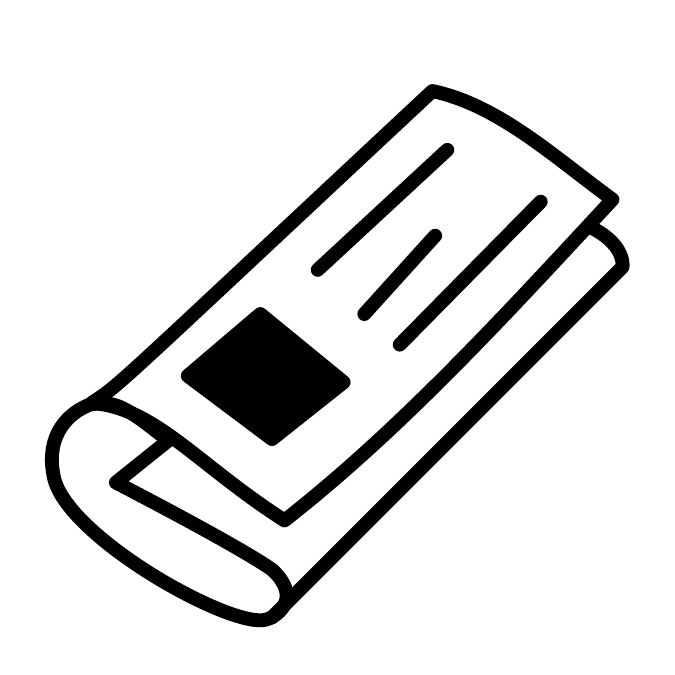} \end{minipage}
        & \begin{minipage}{.03\textwidth}\centering \vspace{.1em}
      \includegraphics[width=5mm, height=5mm]{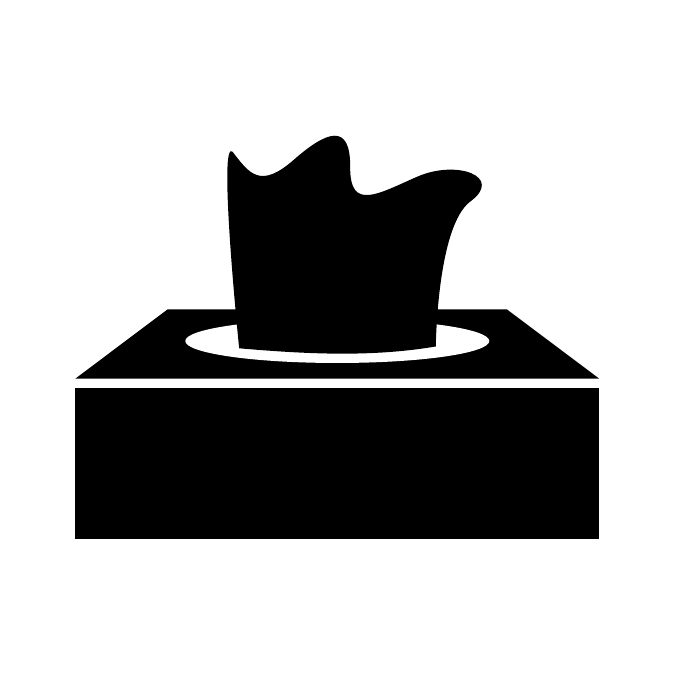} \end{minipage}
        & \begin{minipage}{.03\textwidth}\centering \vspace{.1em}
      \includegraphics[width=5mm, height=5mm]{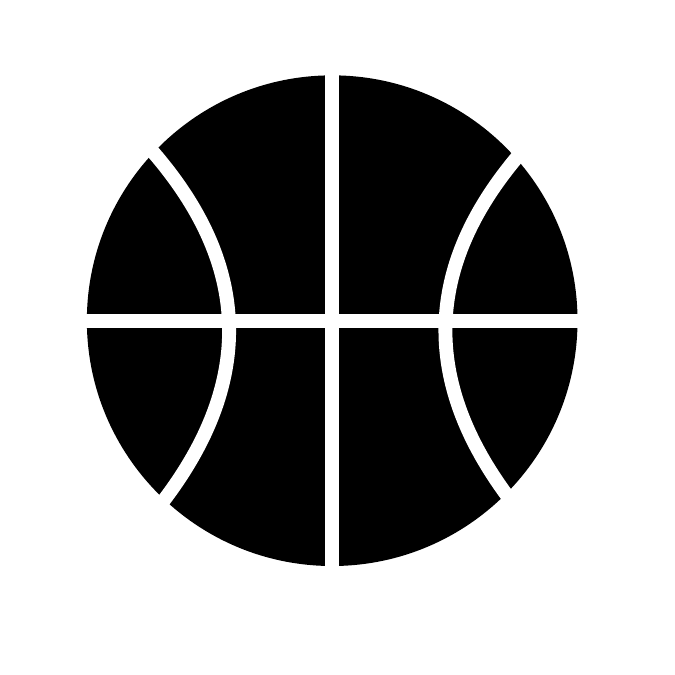} \end{minipage}
        & \begin{minipage}{.03\textwidth}\centering \vspace{.1em}
      \includegraphics[width=5mm, height=5mm]{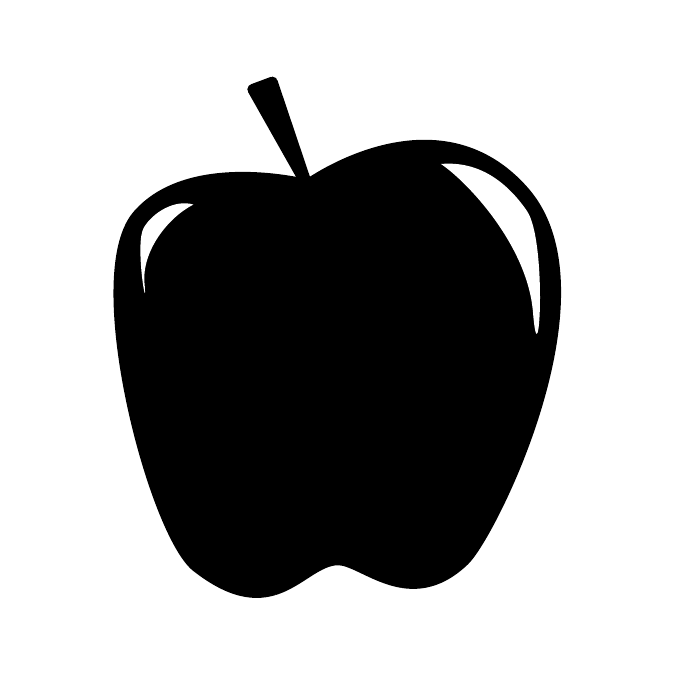} \end{minipage}
        & \begin{minipage}{.03\textwidth}\centering \vspace{.1em}
      \includegraphics[width=5mm, height=5mm]{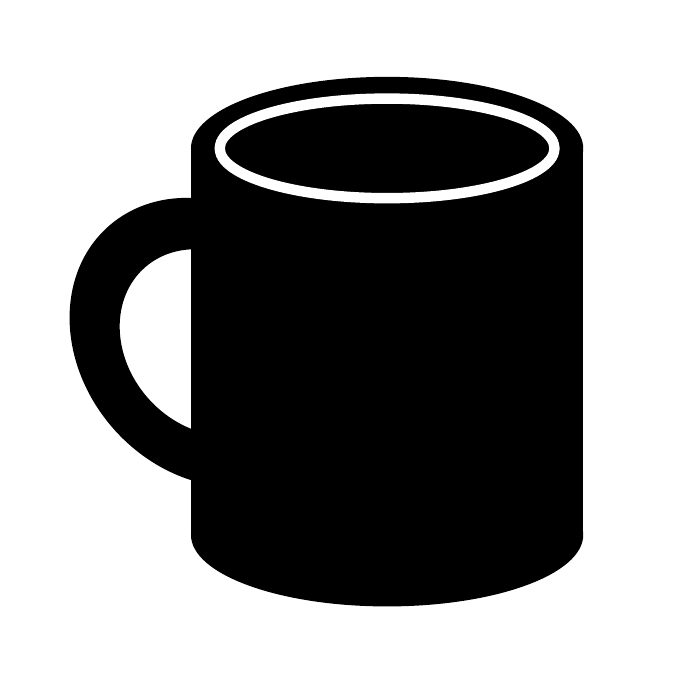} \end{minipage}
        & \begin{minipage}{.03\textwidth}\centering \vspace{.1em}
      \includegraphics[width=5mm, height=5mm]{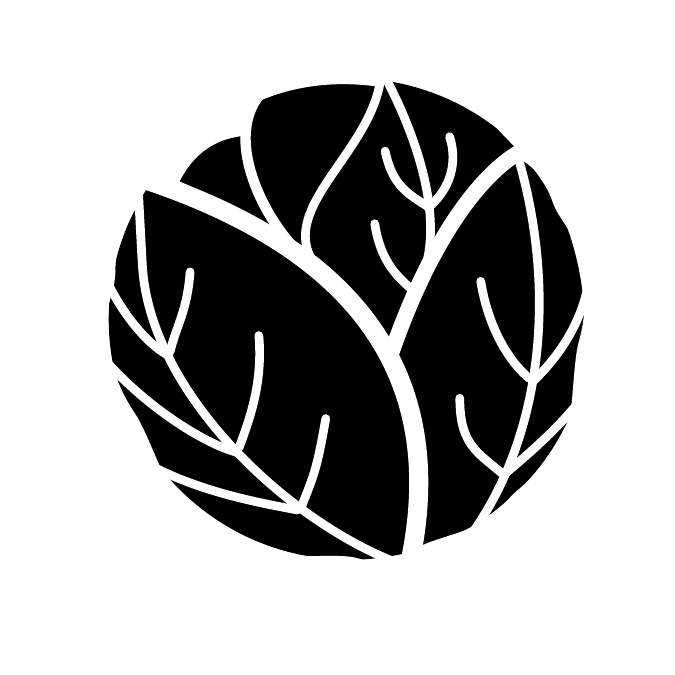} \end{minipage}
        & \begin{minipage}{.03\textwidth}\centering \vspace{.1em}
      \includegraphics[width=5mm, height=5mm]{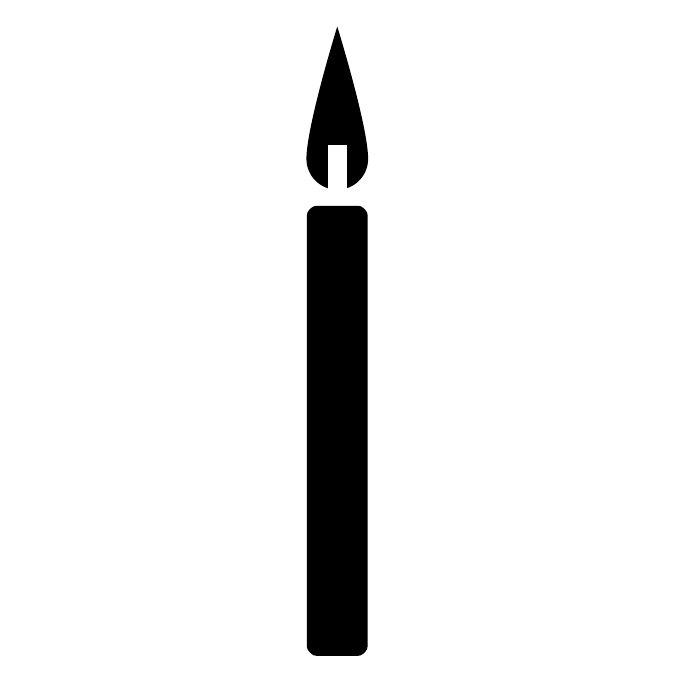} \end{minipage}
      \\ 
	    \hline
        \multicolumn{1}{|c||}{\textbf{Ours, uniform AS}}& 20.4 & 49.8 & 32.4 & 13.7 & 39.2 & 17.3 & 12.9 & 6.0 & 37.6 & 26.8 & 0.0 & 49.6 & 34.8 & 24.0 & 25.6 & 61.2 & 14.0 & 12.8 & 39.6 & 8.4\\ \hline
        \multicolumn{1}{|c||}{\textbf{Ours, full}}& 22.8 & 65.9 & 35.2 & 20.1 & 37.2 & 18.5 & 14.5 & 12.0 & 42.8 & 29.6 & 0.0 & 54.4 & 37.2 & 24.4 & 26.4 & 64.4 & 18.1 & 18.0 & 40.8 & 10.0 \\ \hline 
	\end{tabular}
	\end{minipage}}
	\vspace{-1mm}
	\caption{\textbf{Per category result.} Our dataset includes $20$ object categories. We provide the success rate for each object category.}
	\label{tab:result_category}
	\vspace{-4mm}
\end{table*}

\subsection{Baselines} 
\noindent \textbf{Current Position Predictor (CPP).} This baseline predicts the current position of the object relative to the initial position of the drone in the 3D space, $o_t$, instead of forecasting the future trajectory. The model-predictive planner receives this predicted position at each time-step and outputs the best action for the drone accordingly. The prediction model is trained by an L1 loss with the same training strategy used for our method. 

\noindent \textbf{CPP + Kalman filter.} We implement this baseline by introducing the prediction update through time to the Current Position Predictor (CPP) baseline. We assume the change in the position of the object is linear and follows the Markov assumption in a small time period. Thus, we add the Kalman Filter \cite{welch1995introduction} right after the output of the CPP. To get the transition probability, we average the displacements along the three dimensions over all the trajectories in the training set. We set the process variance to the standard deviation of the average displacements, and measurement variance to $3 \times 10^{-2}$. Further, same as CPP, the model-predictive planner receives this predicted position at each time-step as input and outputs the best action to control the agent. This baseline is expected to be better than CPP, because the Kalman Filter takes into account the possible transitions obtained from the training set so it further smooths out the noisy estimations.


\noindent \textbf{Model-free (A3C \cite{mnih2016asynchronous}).} Another baseline is model-free RL. We use A3C \cite{mnih2016asynchronous} as our model-free RL baseline. The network architecture for A3C includes the same CNN and MLP used in our forecaster and the action sampler. The network receives images $i_{t-2:t}$ as input and directly outputs action $a_t$ for each time-step. We train A3C by $4$ threads and use SharedAdam optimizer with the learning rate of $7\times10^{-4}$. We run the training for $8\times10^{4}$ iterations ($\approx12$ millions frames in total). In addition to using the the `success' signal as the reward, we use the distance between the drone and the object as another reward signal. 

\subsection{Ablations}
We use the training loss described in Sec.~\ref{sec:fn} and the training strategy mentioned in Sec.~\ref{sec:TEdetail} for ablation studies.

\noindent \textbf{Motion Equation (ME).}
The forecaster predicts the position, velocity, and acceleration at the first time-step so we can directly apply motion equation to forecast all future positions. However, since our environment implements complex physical interactions, there are several different types of trajectories (e.g., bouncing or collision). We evaluate if simply using the motion equation is sufficient for capturing such complex behavior.

\noindent \textbf{Uniform Action Sampling (AS).} In this ablation study, we replace our action sampler with a sampler that samples actions from a uniform distribution. This ablation shows the effectiveness of learning a sampler in our model.

\subsection{Results}

\noindent \textbf{Quantitative results.} The results are summarized in Tab.~\ref{tab:result_mo} for all 20 objects and different number of action sequences. We use success rate as our evaluation metric. Recall that the action sampler samples $N$ sequences of future actions. We report results for five different values $N=10,100,1000,10000,100000$. We set the horizon $H$ to $3$ for the forecaster and the planner. For evaluation on the test set, we consider $5K$ episodes for each model. For Tab.~\ref{tab:result_mo}, we repeat the experiments $3$ times and report the average. 

As shown in the table, both the current position predictors (CPP) and the Kalman Filter (CPP + Kalman Filter) baseline are outperformed by our model, which shows the effectiveness of forecasting compared to estimating the current position. Our full method outperforms the model-free baseline, which shows the model-based portion of the model helps improving the performance. `Ours, ME, uniform AS' is worse than the two other variations of our method. This shows that simply applying motion equation and ignoring complex physical interactions is insufficient and it confirms that learning from visual data is necessary. We also show that sampling from a learned policy `Ours - full' outperforms `Ours, uniform AS', which samples from a uniform distribution. This justifies using a learned action sampler and shows the effectiveness of the integration of model-free and model-based learning by the model-predictive planner. 

\subsection{Analysis}\label{sec:analysis}


\begin{figure*}[tp]
    \vspace{-0.3cm}
    \centering
    \includegraphics[width=40pc]{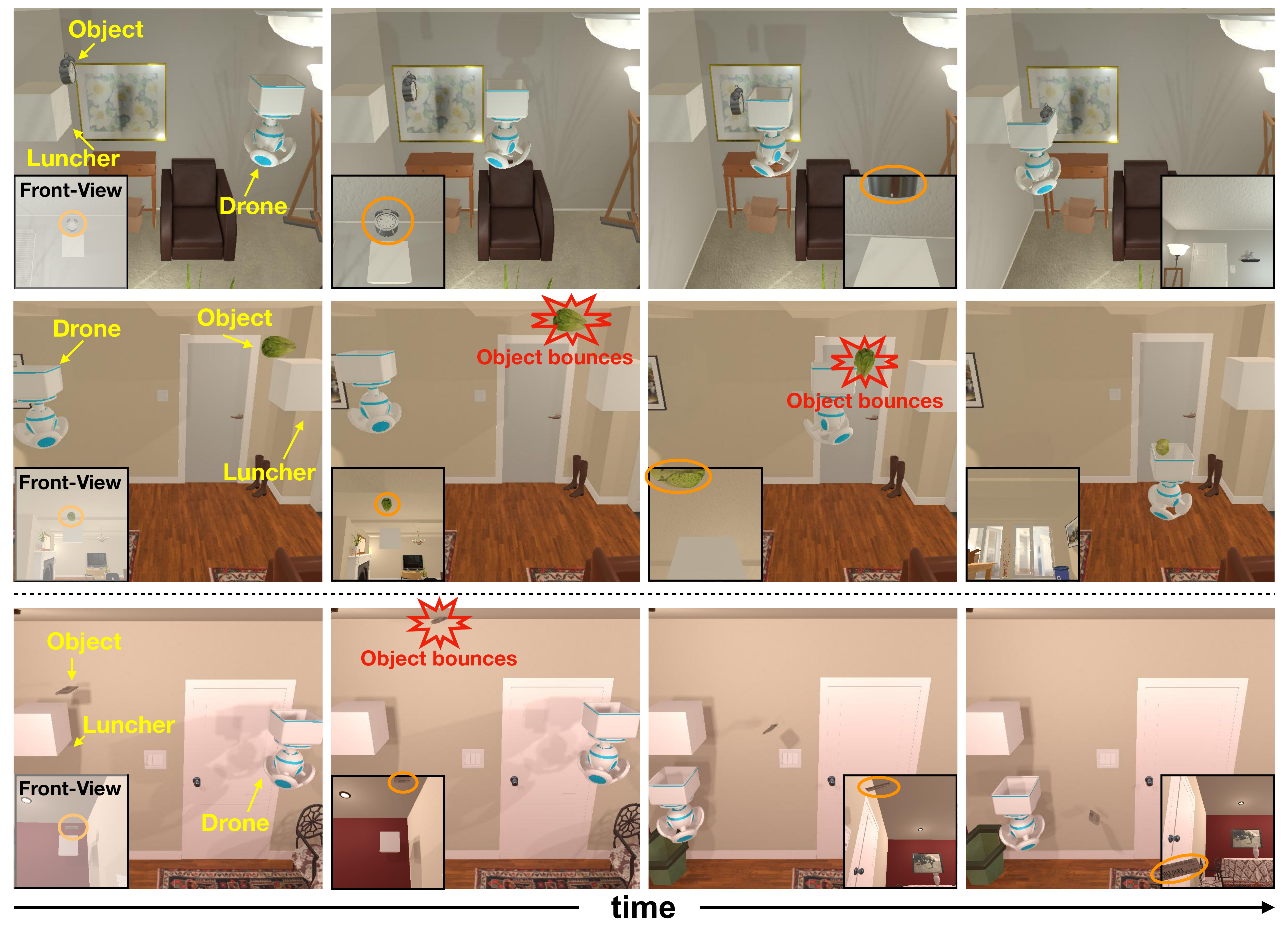}
    \vspace{-4mm}
    \caption{\textbf{Qualitative Results.} We show two successful sequences of catching objects in the first two rows and a failure case in the third row. For instance, in the second row, the object bounces off the ceiling, but the drone is still able to catch it.}
    \label{fig:qualresult}
    \vspace{-5mm}
\end{figure*}

\noindent \textbf{Per-category results.}
Tab.~\ref{tab:result_category} shows the results for each category for `Ours - full' and `Ours, uniform AS'. The results show that our model performs better on relatively heavy objects. This is expected since typically there is less variation in the trajectories of heavy objects.

\noindent \textbf{Difficulty-based categorization.}
Tab.~\ref{tab:result_Difficulty} shows the performance achieved by `Ours - full' and `Ours, uniform AS' in terms of difficulty of the trajectory. The difficulty is defined by how many times the object collides with other structures before reaching the ground or being caught by the agent. We define \emph{easy} by no collision, \emph{medium} by colliding once, and \emph{difficult} by more than one collision. The result shows that even though our model outperforms baselines significantly, it is still not as effective for \emph{medium} and \emph{difficult} trajectories. It suggests that focusing on modeling more complex physical interactions is important for future research.

\begin{table}[tp]
    \setlength\extrarowheight{1pt}
    \setlength{\tabcolsep}{3pt}
	\centering
	\begin{tabular}{c || c | c | c |}
	    \cline{2-4}
	    & \textbf{\ \ Easy\ \ }  & \textbf{Medium} & \textbf{Difficult} \\ 
	    \hline
	    \multicolumn{1}{|c||}{\textbf{Proportion}}& 43$\%$ & 33$\%$ & 24$\%$ \\ \hline
        \multicolumn{1}{|c||}{\textbf{Ours, uniform AS}}& 46.4 & 18.4 & 1.2 \\ \hline
        \multicolumn{1}{|c||}{\textbf{Ours, full}}& 51.9 & 20.7 & 1.6 \\
	    \hline
	\end{tabular}
	\vspace{-1mm}
	\caption{\textbf{Difficulty categorization.} We show the categorization of the results for different levels of difficulty.}
	\label{tab:result_Difficulty}
	\vspace{-3mm}
\end{table}

\noindent \textbf{Different mobility.}
We evaluate how varying the mobility of the drone affects the performance (Tab.~\ref{tab:result_mobility}). We define the mobility as the maximum acceleration of the drone. We re-train the model using $100\%$, $80\%$, $60\%$, $40\%$, $20\%$ of the maximum acceleration. 
\begin{table}[tp]
    \setlength\extrarowheight{1pt}
    \setlength{\tabcolsep}{3pt}
	\centering
	\begin{tabular}{c || c | c | c | c | c |}
	    \cline{2-6}
	    & \textbf{100\%}  & \textbf{80\%} & \textbf{60\%} & \textbf{40\%} & \textbf{20\%} \\ 
	    \hline
        \multicolumn{1}{|c||}{\textbf{Ours, uniform AS}}& 26.0 & 23.6 & 16.0 & 10.5 & 3.3 \\ \hline
        \multicolumn{1}{|c||}{\textbf{Ours, full}}& 29.3 & 25.1 & 18.4 & 10.5 & 3.5 \\
	    \hline
	\end{tabular}
	\vspace{-1mm}
	\caption{\textbf{Mobility results.} We show the results using $100\%$, $80\%$, $60\%$, $40\%$, $20\%$ of the maximum acceleration. }
	\label{tab:result_mobility}
	\vspace{-7mm}
\end{table}



\begin{figure}[!h]
\vspace{-2mm}
    \centering
    \includegraphics[width=20pc,height=11pc]{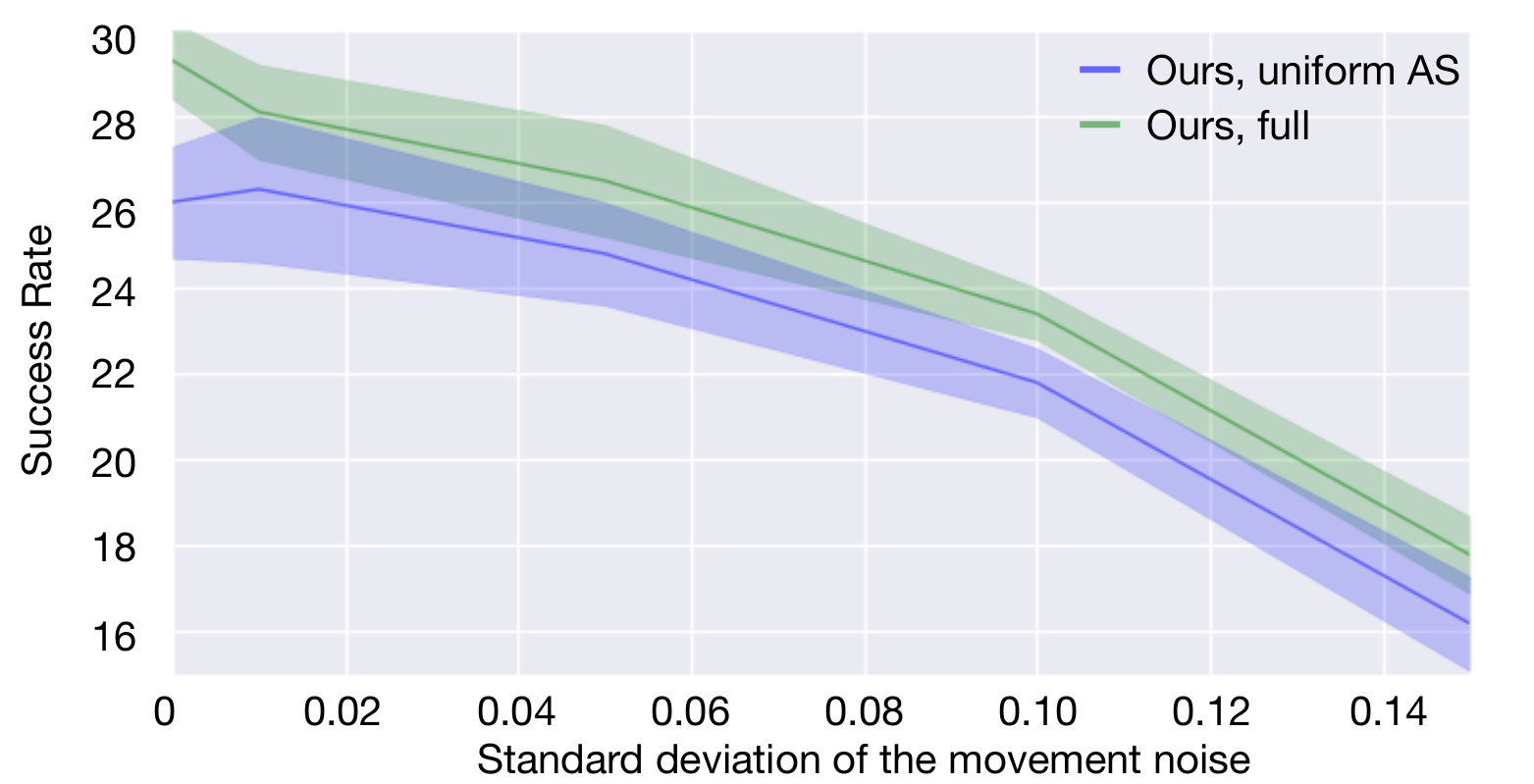} 
    \vspace{-7mm}
    \caption{\textbf{Movement noise variation results.} We show how the noise in the agent movement affects the performance.}
    \label{fig:noise}
    \vspace{-7mm}
\end{figure}

\noindent \textbf{Movement noise.}
Here, we evaluate the scenarios where the agent has more noisy movements. We perform this by injecting a Gaussian noise to the drone's movement after each action (Fig.~\ref{fig:noise}). We re-train the model using $0.01$, $0.05$, $0.1$, and $0.15$ of the standard deviation of the Gaussian noise. As expected, the performance decreases with more noise.


\noindent \textbf{Unseen categories.}
We train the best model on $15$ object categories (the list is in Sec.~\ref{app:E}) and evaluate on the remaining categories. The success rate is $29.12\pm{0.9}$\%. This shows that the model is rather robust to unseen categories.

\noindent \textbf{Qualitative results.} Fig.~\ref{fig:qualresult} shows two sequences of catching
the object and a failure case. The sequence is shown from a third person’s view and the agent camera view (we only use the camera view as the input to our model). The second row shows the drone is still able to catch the object although there is a sudden change in the direction due to the collision of the object with the ceiling. A supplementary video\footnote{\url{https://youtu.be/iyAoPuHxvYs}} shows more success and failure cases.



More analysis are provided in Sec.~\ref{app:F} and Sec.~\ref{app:G}.
\vspace{-2mm}
\section{Conclusion}
\vspace{-1mm}
We address the problem of \textit{visual reaction} in an interactive and dynamic environment in the context of learning to play catch with a drone. This requies learning to forecast the trajectory of the object and to estimate a sequence of actions to intercept the object before it hits the ground. We propose a new dataset for this task, which is built upon the AI2-THOR framework. We showed that the proposed solution outperforms various baselines and ablations of the model including the variations that do not use forecasting, or do not learn a policy based on the forecasting.

\noindent\textbf{Acknowledgements.} We would like to thank Matt Wallingford for valuable feedback and Winson Han and Eli VanderBilt for the design of the drone. This work is in part supported by NSF IIS 1652052, IIS 17303166, DARPA N66001-19-2-4031, 67102239 and gifts from  Allen Institute for Artificial Intelligence.

{\small
\bibliographystyle{ieee_fullname}
\bibliography{egbib}

\begin{thebibliography}{10}\itemsep=-1pt

\bibitem{alahi16}
Alexandre Alahi, Kratarth Goel, Vignesh Ramanathan, Alexandre Robicquet, Li
  Fei-Fei, and Silvio Savarese.
\newblock Social lstm: Human trajectory prediction in crowded spaces.
\newblock In {\em CVPR}, 2016.

\bibitem{bertinetto16}
Luca Bertinetto, Jack Valmadre, Jo{\~a}o~F. Henriques, Andrea Vedaldi, and
  Philip H.~S. Torr.
\newblock Fully-convolutional siamese networks for object tracking.
\newblock In {\em ECCV}, 2016.

\bibitem{bubic2010prediction}
Andreja Bubic, D~Yves Von~Cramon, and Ricarda~I Schubotz.
\newblock Prediction, cognition and the brain.
\newblock {\em Frontiers in human neuroscience}, 4:25, 2010.

\bibitem{Buesing2018LearningAQ}
Lars Buesing, Th{\'e}ophane Weber, S{\'e}bastien Racani{\`e}re, S.~M.~Ali
  Eslami, Danilo~Jimenez Rezende, David~P. Reichert, Fabio Viola, Frederic
  Besse, Karol Gregor, Demis Hassabis, and Daan Wierstra.
\newblock Learning and querying fast generative models for reinforcement
  learning.
\newblock {\em arXiv}, 2018.

\bibitem{castrejon19}
Llu{\'{\i}}s Castrej{\'{o}}n, Nicolas Ballas, and Aaron~C. Courville.
\newblock Improved vrnns for video prediction.
\newblock In {\em ICCV}, 2019.

\bibitem{chao17}
Yu-Wei Chao, Jimei Yang, Brian~L. Price, Scott Cohen, and Jia Deng.
\newblock Forecasting human dynamics from static images.
\newblock In {\em CVPR}, 2017.

\bibitem{chebotar17}
Yevgen Chebotar, Karol Hausman, Marvin Zhang, Gaurav Sukhatme, Stefan Schaal,
  and Sergey Levine.
\newblock Combining model-based and model-free updates for trajectory-centric
  reinforcement learning.
\newblock In {\em ICML}, 2017.

\bibitem{clark2013whatever}
Andy Clark.
\newblock Whatever next? predictive brains, situated agents, and the future of
  cognitive science.
\newblock {\em Behavioral and brain sciences}, 36(3):181--204, 2013.

\bibitem{danelljan17}
Martin Danelljan, Goutam Bhat, Fahad~Shahbaz Khan, and Michael Felsberg.
\newblock Eco: Efficient convolution operators for tracking.
\newblock In {\em CVPR}, 2017.

\bibitem{dosovitskiy17}
Alexey Dosovitskiy and Vladlen Koltun.
\newblock Learning to act by predicting the future.
\newblock In {\em ICLR}, 2017.

\bibitem{feichtenhofer17}
Christoph Feichtenhofer, Axel Pinz, and Andrew Zisserman.
\newblock Detect to track and track to detect.
\newblock In {\em ICCV}, 2017.

\bibitem{feinberg2018model}
V Feinberg, A Wan, I Stoica, MI Jordan, JE Gonzalez, and S Levine.
\newblock Model-based value expansion for efficient model-free reinforcement
  learning.
\newblock In {\em ICML}, 2018.

\bibitem{finn16}
Chelsea Finn, Ian Goodfellow, and Sergey Levine.
\newblock Unsupervised learning for physical interaction through video
  prediction.
\newblock In {\em NeurIPS}, 2016.

\bibitem{fragkiadaki15}
Katerina Fragkiadaki, Sergey Levine, Panna Felsen, and Jitendra Malik.
\newblock Recurrent network models for human dynamics.
\newblock In {\em ICCV}, 2015.

\bibitem{gandhi17}
Dhiraj Gandhi, Lerrel Pinto, and Abhinav Gupta.
\newblock Learning to fly by crashing.
\newblock In {\em IROS}, 2017.

\bibitem{gu16}
Shixiang Gu, Timothy Lillicrap, Ilya Sutskever, and Sergey Levine.
\newblock Continuous deep q-learning with model-based acceleration.
\newblock In {\em ICML}, 2016.

\bibitem{gupta17}
Saurabh Gupta, James Davidson, Sergey Levine, Rahul Sukthankar, and Jitendra
  Malik.
\newblock Cognitive mapping and planning for visual navigation.
\newblock In {\em CVPR}, 2017.

\bibitem{hafner2018planet}
Danijar Hafner, Timothy Lillicrap, Ian Fischer, Ruben Villegas, David Ha,
  Honglak Lee, and James Davidson.
\newblock Learning latent dynamics for planning from pixels.
\newblock {\em arXiv}, 2018.

\bibitem{heess15}
Nicolas Heess, Gregory Wayne, David Silver, Timothy Lillicrap, Tom Erez, and
  Yuval Tassa.
\newblock Learning continuous control policies by stochastic value gradients.
\newblock In {\em NeurIPS}, 2015.

\bibitem{kim14}
Seungsu Kim, Ashwini Shukla, and Aude Billard.
\newblock Catching objects in flight.
\newblock {\em IEEE Transactions on Robotics}, 30:1049--1065, 2014.

\bibitem{kingma2014adam}
Diederik~P Kingma and Jimmy Ba.
\newblock Adam: A method for stochastic optimization.
\newblock {\em arXiv}, 2014.

\bibitem{kitani12}
Kris~M. Kitani, Brian~D. Ziebart, James~Andrew Bagnell, and Martial Hebert.
\newblock Activity forecasting.
\newblock In {\em ECCV}, 2012.

\bibitem{ai2thor}
Eric Kolve, Roozbeh Mottaghi, Winson Han, Eli VanderBilt, Luca Weihs, Alvaro
  Herrasti, Daniel Gordon, Yuke Zhu, Abhinav Gupta, and Ali Farhadi.
\newblock {AI2-THOR: An Interactive 3D Environment for Visual AI}.
\newblock {\em arXiv}, 2017.

\bibitem{kurutach2018model}
Thanard Kurutach, Ignasi Clavera, Yan Duan, Aviv Tamar, and Pieter Abbeel.
\newblock Model-ensemble trust-region policy optimization.
\newblock {\em ICLR}, 2018.

\bibitem{lan14}
Tian Lan, Tsung-Chuan Chen, and Silvio Savarese.
\newblock A hierarchical representation for future action prediction.
\newblock In {\em ECCV}, 2014.

\bibitem{lee17}
Namhoon Lee, Wongun Choi, Paul Vernaza, Christopher~B. Choy, Philip H.~S. Torr,
  and Manmohan Chandraker.
\newblock {DESIRE:} distant future prediction in dynamic scenes with
  interacting agents.
\newblock In {\em CVPR}, 2017.

\bibitem{lerel16}
Adam Lerer, Sam Gross, and Rob Fergus.
\newblock Learning physical intuition of block towers by example.
\newblock {\em arXiv}, 2016.

\bibitem{mathieu16}
Micha{\"e}l Mathieu, Camille Couprie, and Yann LeCun.
\newblock Deep multi-scale video prediction beyond mean square error.
\newblock In {\em ICLR}, 2016.

\bibitem{mirowski17}
Piotr Mirowski, Razvan Pascanu, Fabio Viola, Hubert Soyer, Andrew~J. Ballard,
  Andrea Banino, Misha Denil, Ross Goroshin, Laurent Sifre, Koray Kavukcuoglu,
  Dharshan Kumaran, and Raia Hadsell.
\newblock Learning to navigate in complex environments.
\newblock In {\em ICLR}, 2017.

\bibitem{mnih2016asynchronous}
Volodymyr Mnih, Adria~Puigdomenech Badia, Mehdi Mirza, Alex Graves, Timothy
  Lillicrap, Tim Harley, David Silver, and Koray Kavukcuoglu.
\newblock Asynchronous methods for deep reinforcement learning.
\newblock In {\em ICML}, 2016.

\bibitem{mottaghi16}
Roozbeh Mottaghi, Hessam Bagherinezhad, Mohammad Rastegari, and Ali Farhadi.
\newblock Newtonian image understanding: Unfolding the dynamics of objects in
  static images.
\newblock In {\em CVPR}, 2016.

\bibitem{mottaghi16b}
Roozbeh Mottaghi, Mohammad Rastegari, Abhinav Gupta, and Ali Farhadi.
\newblock "what happens if..." learning to predict the effect of forces in
  images.
\newblock In {\em ECCV}, 2016.

\bibitem{muller11}
Mark Muller, Sergei Lupashin, and Raffaello D'Andrea.
\newblock Quadrocopter ball juggling.
\newblock In {\em IROS}, 2011.

\bibitem{nagabandi18}
Anusha Nagabandi, Gregory Kahn, Ronald~S. Fearing, and Sergey Levine.
\newblock Neural network dynamics for model-based deep reinforcement learning
  with model-free fine-tuning.
\newblock In {\em ICRA}, 2018.

\bibitem{nam16}
Hyeonseob Nam and Bohyung Han.
\newblock Learning multi-domain convolutional neural networks for visual
  tracking.
\newblock In {\em CVPR}, 2016.

\bibitem{park16}
Hyun~Soo Park, Jyh-Jing Hwang, Yedong Niu, and Jianbo Shi.
\newblock Egocentric future localization.
\newblock In {\em CVPR}, 2016.

\bibitem{pintea14}
Silvia~L. Pintea, Jan~C. van Gemert, and Arnold W.~M. Smeulders.
\newblock D{\'e}j{\`a} vu: Motion prediction in static images.
\newblock In {\em ECCV}, 2014.

\bibitem{racaniere17}
S\'{e}bastien Racani\`{e}re, Theophane Weber, David Reichert, Lars Buesing,
  Arthur Guez, Danilo Jimenez~Rezende, Adri\`{a} Puigdom\`{e}nech~Badia, Oriol
  Vinyals, Nicolas Heess, Yujia Li, Razvan Pascanu, Peter Battaglia, Demis
  Hassabis, David Silver, and Daan Wierstra.
\newblock Imagination-augmented agents for deep reinforcement learning.
\newblock In {\em NeurIPS}, 2017.

\bibitem{rhinehart17}
Nicholas Rhinehart and Kris~M. Kitani.
\newblock First-person activity forecasting with online inverse reinforcement
  learning.
\newblock In {\em ICCV}, 2017.

\bibitem{ritz12}
Robin Ritz, Mark~W. M{\"u}ller, Markus Hehn, and Raffaello D'Andrea.
\newblock Cooperative quadrocopter ball throwing and catching.
\newblock In {\em IROS}, 2012.

\bibitem{sadeghi17}
Fereshteh Sadeghi and Sergey Levine.
\newblock {CAD2RL:} real single-image flight without a single real image.
\newblock In {\em RSS}, 2017.

\bibitem{sandler2018mobilenetv2}
Mark Sandler, Andrew Howard, Menglong Zhu, Andrey Zhmoginov, and Liang-Chieh
  Chen.
\newblock Mobilenetv2: Inverted residuals and linear bottlenecks.
\newblock In {\em CVPR}, 2018.

\bibitem{savinov18}
Nikolay Savinov, Alexey Dosovitskiy, and Vladlen Koltun.
\newblock Semi-parametric topological memory for navigation.
\newblock In {\em ICLR}, 2018.

\bibitem{silva15}
Rui Silva, Francisco~S. Melo, and Manuela~M. Veloso.
\newblock Towards table tennis with a quadrotor autonomous learning robot and
  onboard vision.
\newblock In {\em IROS}, 2015.

\bibitem{silver17}
David Silver, Hado van Hasselt, Matteo Hessel, Tom Schaul, Arthur Guez, Tim
  Harley, Gabriel Dulac-Arnold, David Reichert, Neil Rabinowitz, Andre Barreto,
  and Thomas Degris.
\newblock The predictron: End-to-end learning and planning.
\newblock In {\em ICML}, 2017.

\bibitem{srivastava15}
Nitish Srivastava, Elman Mansimov, and Ruslan Salakhutdinov.
\newblock Unsupervised learning of video representations using lstms.
\newblock In {\em ICML}, 2015.

\bibitem{su17}
Kunyue Su and Shaojie Shen.
\newblock Catching a flying ball with a vision-based quadrotor.
\newblock In {\em ISER}, 2017.

\bibitem{sun18}
Chong Sun, Huchuan Lu, and Ming-Hsuan Yang.
\newblock Learning spatial-aware regressions for visual tracking.
\newblock In {\em CVPR}, 2018.

\bibitem{chen19}
Chen Sun, Abhinav Shrivastava, Carl Vondrick, Rahul Sukthankar, Kevin Murphy,
  and Cordelia Schmid.
\newblock Relational action forecasting.
\newblock In {\em CVPR}, 2019.

\bibitem{sutton2000policy}
Richard~S Sutton, David~A McAllester, Satinder~P Singh, and Yishay Mansour.
\newblock Policy gradient methods for reinforcement learning with function
  approximation.
\newblock In {\em NeurIPS}, 2000.

\bibitem{tamar16}
Aviv Tamar, Yi Wu, Garrett Thomas, Sergey Levine, and Pieter Abbeel.
\newblock Value iteration networks.
\newblock In {\em NeurIPS}, 2016.

\bibitem{villegas19}
Ruben Villegas, Arkanath Pathak, Harini Kannan, Dumitru Erhan, Quoc~V. Le, and
  Honglak Lee.
\newblock High fidelity video prediction with large stochastic recurrent neural
  networks.
\newblock In {\em NeurIPS}, 2019.

\bibitem{villegas17}
Ruben Villegas, Jimei Yang, Seunghoon Hong, Xunyu Lin, and Honglak Lee.
\newblock Decomposing motion and content for natural video sequence prediction.
\newblock In {\em ICLR}, 2017.

\bibitem{vondrick16}
Carl Vondrick, Hamed Pirsiavash, and Antonio Torralba.
\newblock Anticipating visual representations from unlabeled video.
\newblock In {\em CVPR}, 2016.

\bibitem{walker16}
Jacob Walker, Carl Doersch, Abhinav Gupta, and Martial Hebert.
\newblock An uncertain future: Forecasting from static images using variational
  autoencoders.
\newblock In {\em ECCV}, 2016.

\bibitem{walker14}
Jacob Walker, Abhinav Gupta, and Martial Hebert.
\newblock Patch to the future: Unsupervised visual prediction.
\newblock In {\em CVPR}, 2014.

\bibitem{walker15}
Jacob Walker, Abhinav Gupta, and Martial Hebert.
\newblock Dense optical flow prediction from a static image.
\newblock In {\em ICCV}, 2015.

\bibitem{walke17}
Jacob Walker, Kenneth Marino, Abhinav Gupta, and Martial Hebert.
\newblock The pose knows: Video forecasting by generating pose futures.
\newblock In {\em ICCV}, 2017.

\bibitem{wang18}
Xin Wang, Wenhan Xiong, Hongmin Wang, and William~Yang Wang.
\newblock Look before you leap: Bridging model-free and model-based
  reinforcement learning for planned-ahead vision-and-language navigation.
\newblock In {\em ECCV}, 2018.

\bibitem{watters17}
Nicholas Watters, Daniel Zoran, Theophane Weber, Peter~W. Battaglia, Razvan
  Pascanu, and Andrea Tacchetti.
\newblock Visual interaction networks: Learning a physics simulator from video.
\newblock In {\em NeurIPS}, 2017.

\bibitem{welch1995introduction}
Greg Welch, Gary Bishop, et~al.
\newblock An introduction to the kalman filter.
\newblock 1995.

\bibitem{xie13}
Dan Xie, Sinisa Todorovic, and Song-Chun Zhu.
\newblock Inferring "dark matter" and "dark energy" from videos.
\newblock In {\em ICCV}, 2013.

\bibitem{xue16}
Tianfan Xue, Jiajun Wu, Katherine Bouman, and Bill Freeman.
\newblock Visual dynamics: Probabilistic future frame synthesis via cross
  convolutional networks.
\newblock In {\em NeurIPS}, 2016.

\bibitem{yang19}
Wei Yang, Xiaolong Wang, Ali Farhadi, Abhinav Gupta, and Roozbeh Mottaghi.
\newblock Visual semantic navigation using scene priors.
\newblock In {\em ICLR}, 2019.

\bibitem{yuen10}
Jenny Yuen and Antonio Torralba.
\newblock A data-driven approach for event prediction.
\newblock In {\em ECCV}, 2010.

\bibitem{zheng14}
Bo Zheng, Yibiao Zhao, Joey~C. Yu, Katsushi Ikeuchi, and Song-Chun Zhu.
\newblock Scene understanding by reasoning stability and safety.
\newblock {\em IJCV}, 2014.

\bibitem{zhu17}
Yuke Zhu, Roozbeh Mottaghi, Eric Kolve, Joseph~J. Lim, Abhinav Gupta, Li
  Fei-Fei, and Ali Farhadi.
\newblock Target-driven visual navigation in indoor scenes using deep
  reinforcement learning.
\newblock In {\em ICRA}, 2017.

\end{thebibliography}
}

\clearpage
\vfill

\appendix
\twocolumn[{%
\renewcommand\twocolumn[1][]{#1}
\begin{center}
    \captionsetup{type=figure}
	\includegraphics[width=40pc,height=9pc]{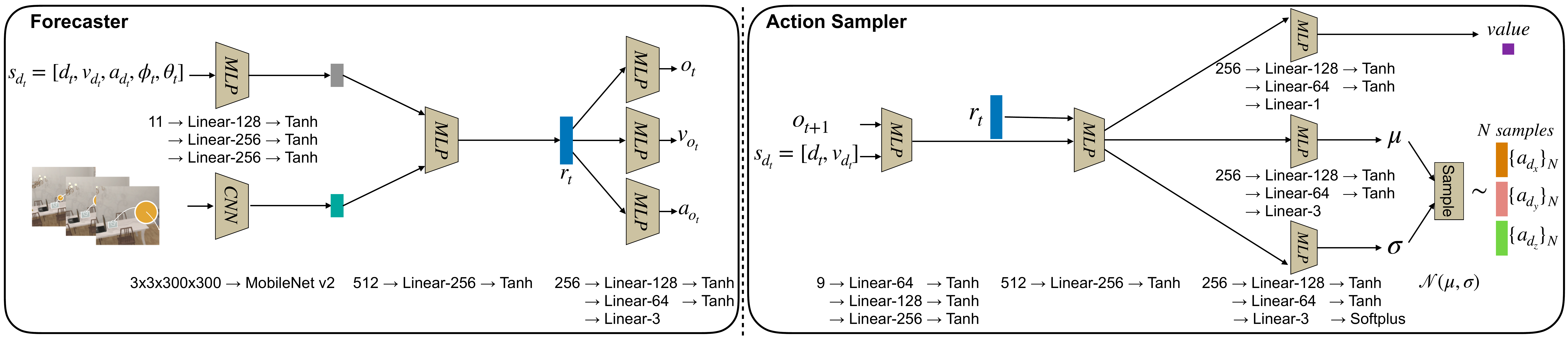}
	\captionof{figure}{\textbf{Detailed architecture of the forecaster and action sampler.}
	}
	\vspace{3mm}
	\label{fig:detail_model}
\end{center}
}]

\section{Complete list of objects}\label{app:A}

We use $20$ objects for the experiments: \textit{alarm clock}, \textit{apple}, \textit{basketball}, \textit{book}, \textit{bowl}, \textit{bread}, \textit{candle}, \textit{cup}, \textit{glass bottle}, \textit{lettuce}, \textit{mug}, \textit{newspaper}, \textit{salt shaker}, \textit{soap bottle}, \textit{statue}, \textit{tissue box}, \textit{toaster}, \textit{toilet paper}, \textit{vase} and \textit{watering can}.

\section{Results for the case that the camera is fixed}\label{app:B}

In Tab.~\ref{tab:result_camera}, we provide the results for the case that the drone camera is fixed and does not rotate. In this experiment, we set horizon $H=3$ and number of action sequences $N=100,000$. The performance degrades for the case that the camera does not rotate, which is expected.

\begin{table}[h]
    \setlength\extrarowheight{1pt}
    \setlength{\tabcolsep}{3pt}
	\centering
	\begin{tabular}{c || c | c | c |}
	    \cline{2-4}
	    & \textbf{GT.}  & \textbf{Est.} & \textbf{Fixed}\\ 
	    \hline
        \multicolumn{1}{|c||}{\textbf{Ours, uniform AS}}& 54.7 & 26.0 & 18.6 \\ \hline
        \multicolumn{1}{|c||}{\textbf{Ours, full}}& 59.3 & 29.3 & 20.2 \\
	    \hline
	\end{tabular}
	\caption{\textbf{Camera orientation results.} We show the results for the scenario that the camera orientation does not change. \textbf{GT} corresponds to the case that we use the ground truth camera orientation at train/test time. The ground truth camera orientation is obtained via ground truth object's position and drone's position. \textbf{Est} denotes the case that we use the predicted  object and drone positions to calculate to estimate the camera angle.  \textbf{Fixed} denotes the case that the camera orientation is fixed.}
	\label{tab:result_camera}
\end{table}

\section{More statistics of object properties}\label{app:C}

\begin{figure*}[htp]
    \centering
    \includegraphics[width=40pc,height=30pc]{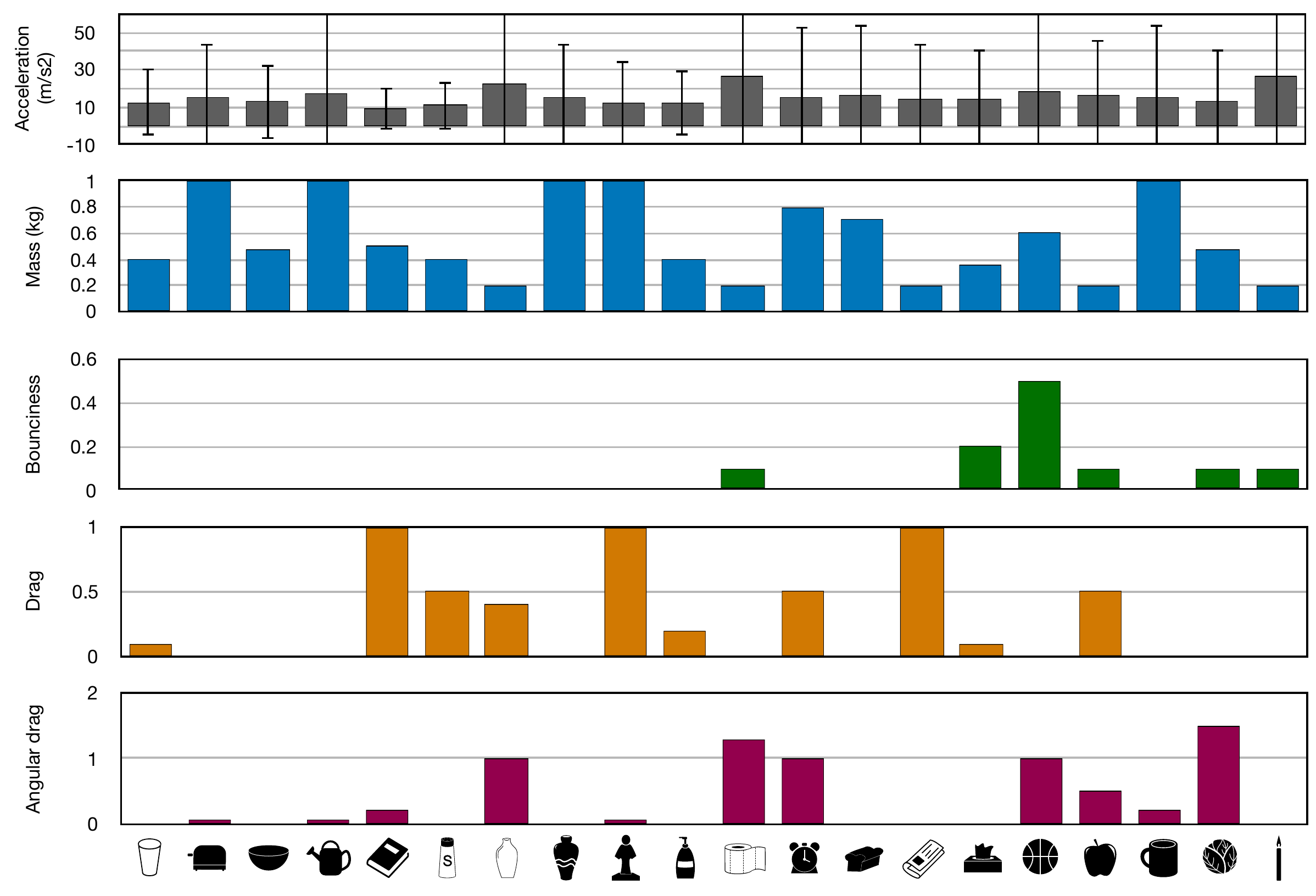} 
    \caption{\textbf{More dataset statistics.} We provide more statistics for the 20 types of objects in our dataset. We illustrate the mass, average acceleration along the trajectories, bounciness, drag, and angular drag.}
    \label{fig:more_stats}
\end{figure*}

We show more statistics about our dataset in the Fig.~\ref{fig:more_stats}, including the \textit{mass}, \textit{average acceleration} along the trajectories, \textit{bounciness}, \textit{drag}, and \textit{angular drag}. Drag is the tendency of an object to slow down due to friction.

\section{Details of the model architecture}\label{app:D}

Fig.~\ref{fig:detail_model} summarizes the details of the model architecture. 

\section{List of objects for the unseen categories experiment}\label{app:E}

We selected a subset of $5$ objects as our held-out set such that they have different physical properties: \textit{basketball},  \textit{bowl}, \textit{bread}, \textit{candle}, \textit{watering can}. We trained our model on the rest of the objects: \textit{alarm clock}, \textit{apple}, \textit{book},  \textit{cup}, \textit{glass bottle}, \textit{lettuce}, \textit{mug}, \textit{newspaper}, \textit{salt shaker}, \textit{soap bottle}, \textit{statue}, \textit{tissue box}, \textit{toaster}, \textit{toilet paper} and \textit{vase}.

\section{Visualization of forecasted trajectory}\label{app:F}

\begin{figure}[h]
    \centering
    \includegraphics[width=21pc]{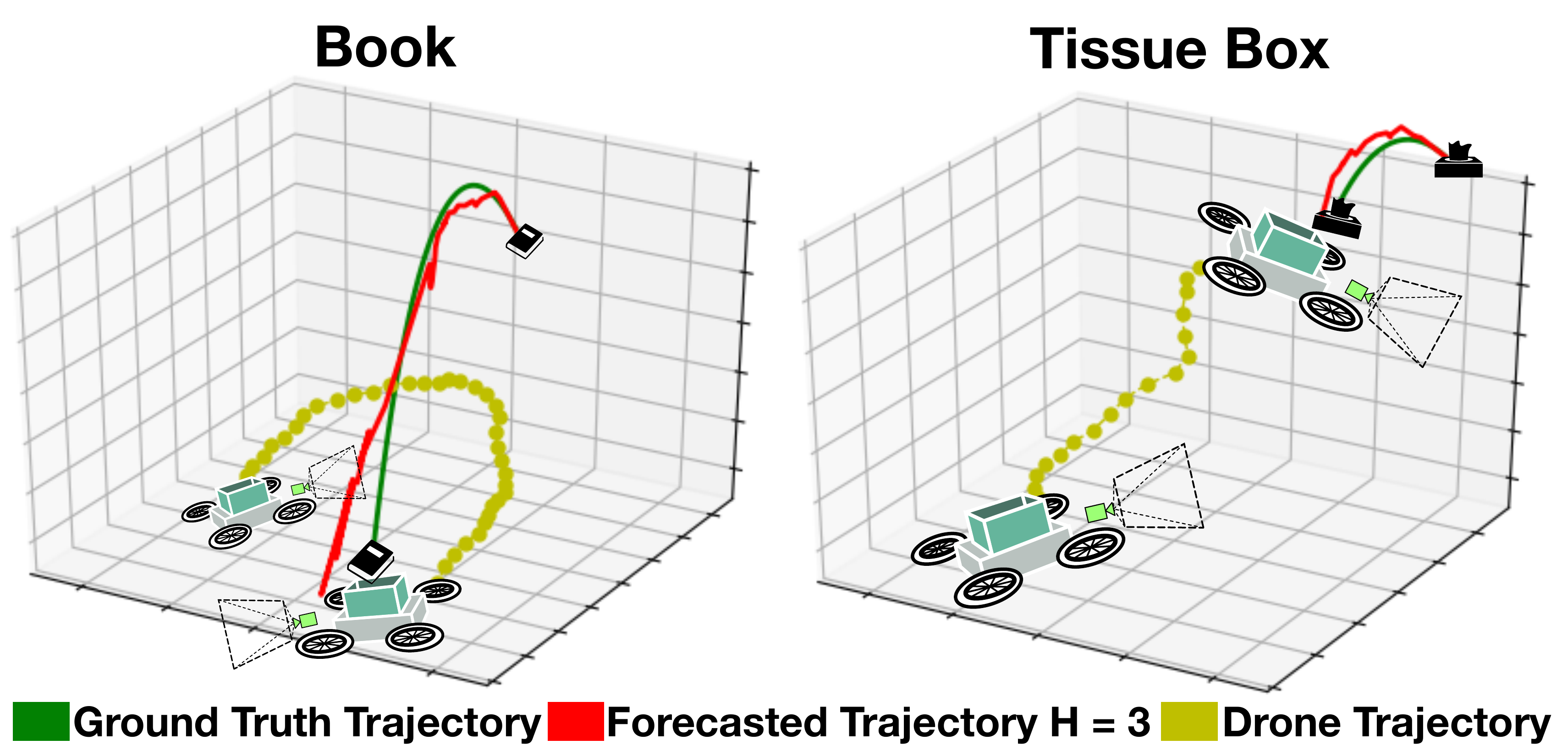}
    \caption{\textbf{Qualitative examples of the forecasting trajectory.} The green color, red color, and yellow color denote the ground truth object's trajectory, forecasted object's trajectory, and drone's moving trajectory, respectively.}
    \label{fig:result}
    \vspace{-7mm}
\end{figure}

We visualize two examples of the forecasted trajectory in Figure~\ref{fig:result}. The examples demonstrate that the drone is able to plan according to the future positions of the objects predicted by the forecaster.

\section{Error of position, velocity and acceleration prediction.}\label{app:G} The error of our method (L2 distance) for predicting position, velocity and acceleration are 0.644$\pm$0.389, 0.037$\pm$0.017, and 0.007$\pm$0.014, respectively. The error for the baseline CPP for example is 0.686$\pm$0.362, 0.148$\pm$0.033, and 0.007$\pm$0.013 for position, velocity and acceleration, respectively. This comparison is performed for $N=100,000$.



\section{Different horizon length}\label{app:H}
\begin{figure}[!h]
    \centering
    \includegraphics[width=20pc,height=11pc]{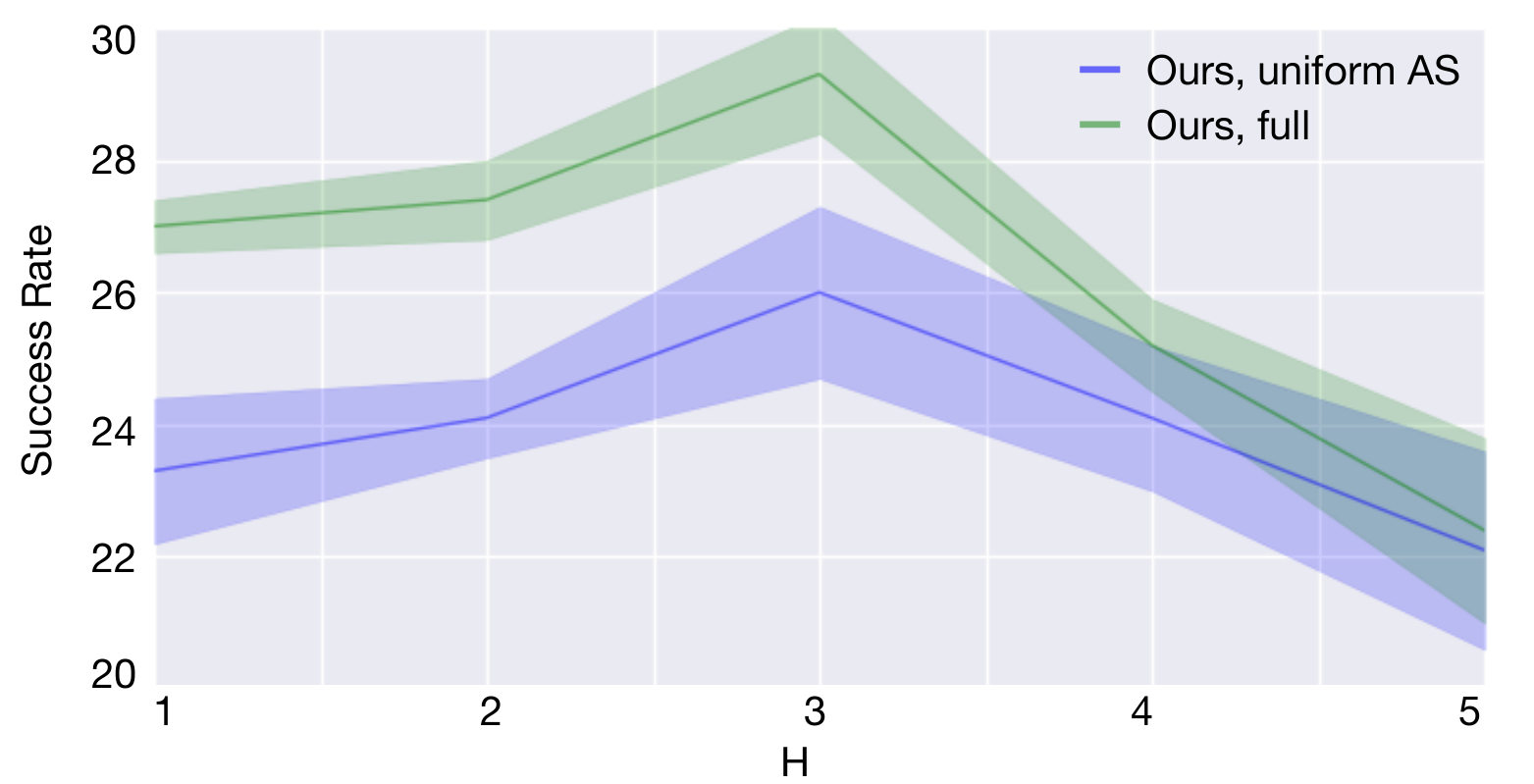} 
    \vspace{-5mm}
    \caption{\textbf{Result for different horizon lengths.} We show how the performance changes by varying $H$.}
    \label{fig:horizon}
    \vspace{-5mm}
\end{figure}

Here, we show how the performance changes with varying the horizon length $H$ (Fig.~\ref{fig:horizon}).
We observe a performance decrease for horizons longer than 3. The reason is that the learned forecaster has a small error and the error for each time-step accumulates. Thus, training an effective model with longer horizons is challenging and we leave it for future research.

\end{document}